\documentclass[10pt,journal,compsoc]{IEEEtran}
\usepackage{times}
\usepackage{epsfig}
\usepackage{graphicx}
\usepackage{amsmath}
\usepackage{amssymb}

\usepackage{subfigure}
\usepackage[section]{algorithm}
\usepackage{algorithmic}

\usepackage{makecell}
\usepackage{multirow}
\usepackage{array}
\usepackage{colortbl}
\usepackage{tikz}
\usepackage{enumitem}
\usepackage{comment}

\newcommand{\cD}{\mathcal{D}}
\newcommand{\cG}{\mathcal{G}}

\newcommand{\cI}{\mathcal{I}}

\newcommand{\cL}{\mathcal{L}}

\newcommand{\R}{\mathbb{R}}

\newcommand{\suml}[2]{\sum\limits_{#1}^{#2}}

\newcommand{\etal}{\emph{et al.}}
\newcommand{\eg}{\emph{e.g.}}
\newcommand{\ie}{\emph{i.e.}}
\newcommand{\vs}{\emph{vs.}}
\newcolumntype{L}[1]{>{\raggedright\let\newline\\\arraybackslash\hspace{0pt}}m{#1}}
\newcolumntype{C}[1]{>{\centering\let\newline\\\arraybackslash\hspace{0pt}}m{#1}}
\newcolumntype{R}[1]{>{\raggedleft\let\newline\\\arraybackslash\hspace{0pt}}m{#1}}

\newcommand{\tikzmark}[1]{\tikz[remember picture,overlay, baseline=-0.5ex]\node (#1){};}
\newcommand{\connect}[3][3mm]{\tikz[remember picture,overlay]\draw[shorten <=-#1, shorten >=-#1](#2)--(#3);}

\definecolor{grayB}{rgb}{0.85,0.85,0.85}
\usepackage[pagebackref=false,breaklinks=true,letterpaper=true,colorlinks,bookmarks=false]{hyperref}
\begin{document}
\title{Trainable Nonlinear Reaction Diffusion: A Flexible Framework for Fast and Effective \\ Image Restoration}
\author{Yunjin~Chen and Thomas~Pock
\IEEEcompsocitemizethanks{\IEEEcompsocthanksitem Y.J. Chen is with the Institute for Computer Graphics and Vision, 
Graz University of Technology, 8010 Graz, Austria. 
\protect\\
E-mail: {chenyunjin\_nudt}@hotmail.com
\IEEEcompsocthanksitem T.~Pock is with the Institute for Computer Graphics and Vision, 
Graz University of Technology, 8010 Graz, Austria, as well as 
Digital Safety \& Security Department, AIT Austrian Institute of Technology 
GmbH, 1220 Vienna, Austria. 
E-mail: pock@icg.tugraz.at \protect\\
This work was supported by the
   Austrian Science Fund (FWF) under the China Scholarship Council
 (CSC) Scholarship Program and the START project BIVISION,
 No. Y729. 
}
}

\markboth{IEEE Transactions on Pattern Analysis and Machine Intelligence ,~Vol.~xx, No.~xx, 2016}%
{Chen \MakeLowercase{\textit{et al.}}: Trainable nonlinear reaction diffusion}

\IEEEtitleabstractindextext{%
\begin{abstract}
Image restoration is a long-standing problem in low-level computer vision with 
many interesting applications. {
We describe a flexible learning framework based on the concept of nonlinear 
reaction diffusion models for various image 
restoration problems. By embodying recent 
improvements in nonlinear diffusion models, we propose a dynamic nonlinear reaction diffusion model with time-dependent parameters (\ie, linear filters 
and influence functions).} 
In contrast to previous nonlinear diffusion models, 
all the parameters, including the filters and the influence functions, 
are simultaneously learned from training data through a loss based approach. 
We call this approach TNRD -- \textit{Trainable Nonlinear Reaction Diffusion}. 
The TNRD approach is applicable for a variety of image restoration tasks 
by incorporating appropriate reaction force. 
We demonstrate its capabilities with three representative applications, Gaussian 
image denoising, single image super resolution and 
JPEG deblocking. Experiments show that our trained nonlinear diffusion models 
largely benefit from the training 
of the parameters and finally lead to the best reported performance on 
common test datasets for the tested applications. 
Our trained models preserve the structural simplicity of diffusion models and 
take only a small number of diffusion steps, thus are highly efficient. 
Moreover, they are also well-suited for parallel computation on GPUs, 
which makes the inference procedure extremely fast.
\end{abstract}

}
\maketitle

\IEEEdisplaynontitleabstractindextext
\IEEEpeerreviewmaketitle

\IEEEraisesectionheading{\section{Introduction}\label{sec:introduction}}
\IEEEPARstart{I}{mage} restoration is the process of estimating uncorrupted images from
noisy or blurred ones. It is one of the most fundamental operations in
image processing, video processing, and low-level computer vision. 
For several decades, image restoration remains an active research topic and hence new 
approaches are constantly emerging. There exists a huge amount of literature addressing the topic of image
restoration problems, see for example \cite{milanfar2013tour} for a survey. 

In recent years, the predominant approaches for image restoration 
are non-local methods based on patch modeling, 
for example, image denoising with (i) Gaussian noise 
\cite{LSSC, WNNM, BM3D, rajwade2013image}, 
(ii) multiplicative noise \cite{CozzolinoPSPV14}, or (iii) 
Poisson noise \cite{Giryes}, 
image interpolation \cite{RomanoPE14}, 
image deconvolution \cite{dong2013nonlocally}, etc. 
Most state-of-the-art techniques mainly
concentrate on achieving utmost image restoration quality, with little 
consideration on the computational efficiency, 
\eg, \cite{LSSC, WNNM, RomanoPE14}, despite the fact that 
it is a critical factor for real applications. However, there are a few exceptions. For example, 
there are two notable exceptions for the task of Gaussian denoising, 
BM3D \cite{BM3D} and the recently proposed Cascade of Shrinkage Fields (CSF) 
\cite{CSF2014} model, which simultaneously offer high efficiency and high image restoration quality.

It is well-known that BM3D is a highly engineered Gaussian image
denoising algorithm. It involves a block matching process, which is
challenging for parallel computation on GPUs, alluding to the fact
that it is not straightforward to accelerate BM3D algorithm on
parallel architectures. In contrast, the recently proposed CSF model
offers high levels of parallelism, making it well suited for GPU
implementation, thus owning high computational efficiency.

In this paper, we propose a flexible learning framework to generate
fast and effective models for a variety of image restoration
problems. Our approach is based on learning optimal nonlinear
reaction diffusion models. The learned models preserve the structural
simplicity of these models and hence it is straightforward to
implement the corresponding algorithms on massive parallel hardware
such as GPUs.

\subsection{Nonlinear diffusion for image restoration}\label{intro} {
  Partial differential equation (PDEs) have become a standard approach
  for various problems in image processing. On the one hand they come
  along with a sound mathematical framework that allow to make clear
  statements about the existence and regularity of the solutions. On
  the other hand, efficient numerical algorithms have been developed,
  that allow to compute the solution of PDEs in very short
  time~\cite{PMmodel, anisotropicbook} . Although recent PDE approaches
  have shown good performance for a number of image processing task,
  they still fail to produce state-of-the-art quality for classical
  image restoration tasks.

In the seminal work \cite{PMmodel}, Perona and Malik (PM) proposed a 
nonlinear diffusion model, which is given as the following PDE
\begin{equation}\label{PM}
\begin{cases}
{\frac{\partial u}{\partial t} = \text{div} (g(|\nabla u|) \nabla u)}\\
u \big|_{t = 0} = f \,, 
\end{cases}
\end{equation}
{where $\nabla$ is the gradient operator, $t$ denotes the time, $f$ is 
  a initial image.} The function $g$ is known as edge-stopping function
\cite{ROBUSTANISOTROPIC} or diffusivity function \cite{anisotropicbook}, and 
a typical $g$-function is given by $g(z) ={1}/{(1+z^2)}$. 
The proposed PM diffusion model \eqref{PM} leads to a 
nonlinear anisotropic\footnote{
  Anisotropic diffusion in this paper is understood in the sense that 
  the smoothing induced by PDEs can be favored in some directions and 
  prevented in others. The diffusivity is not 
  necessary to be a tensor. It should be noted that this definition is different from Weickert's terminology \cite{anisotropicbook}, 
  where anisotropic diffusion always involves a diffusion tensor, and the PM model is regarded as an isotropic model.}
diffusion model which is able to preserve and enhance image edges.
Hence, it is well suited for a number of image processing tasks such as image denoising and segmentation. 

\subsubsection{Improvements from the side of PDEs}\label{sec:pdes}
A first variant of the PM model is the so-called biased anisotropic diffusion 
(also known as reaction diffusion) proposed by Nordstr{\"o}m \cite{biased}, which introduces a bias term (forcing term) to 
free the user from the difficulty of specifying an appropriate stopping time for the PM diffusion process. 
This additional term reacts against the strict smoothing effect of the pure PM diffusion, therefore resulting in a nontrivial steady-state. 

Subsequent works consider modifications of the diffusion or the reaction 
term for the reaction diffusion model 
\cite{esclarin1997image, cottet1993image, acton2001oriented}, \eg, 
Acton \etal ~\cite{acton2001oriented} exploited 
a more complicated reaction term to enhance oriented 
textures; \cite{barcelos2003well} proposed 
to replace the ordinary diffusion term with a flow equation based on mean curvature. 
A notable work is the forward-backward diffusion model proposed by Gilboa \etal~ \cite{gilboa2002forward}, 
which incorporates explicit 
inverse diffusion with negative diffusivity coefficient by carefully choosing the diffusivity function. The resulting diffusion 
processes can adaptively switch between forward and backward diffusion processes. 
In subsequent work \cite{welk2009theoretical}, 
the theoretical framework for discrete forward-and-backward diffusion 
filtering has been investigated. 
Researchers also propose to exploit higher-order nonlinear diffusion filtering, which involves larger linear filters, 
\eg, fourth-order diffusion models \cite{hajiaboli2011anisotropic, didas2009properties, guidotti2011two}.
Meanwhile, theoretical properties about the stability and local feature enhancement of higher-order nonlinear diffusion filtering 
are established in \cite{didas2005stability}. 

It should be noted that the above mentioned diffusion models are 
handcrafted, including elaborate selections of the diffusivity functions, optimal stopping times and proper reaction forces. 
It is a generally difficult task to design a good-performing PDE for 
a specific image processing problem, as good insights into this problem and a deep understanding of the behavior of the PDEs are usually required. 
Therefore, an attempt to learn PDEs from training data via an 
optimal control approach was made in \cite{liu2010learning}, where 
the PDEs to be trained have the form of 
\begin{equation}\label{eq.liu}
\begin{cases}
\frac{\partial u}{\partial t} = \kappa(u) + a(t)^\top \mathcal{O}(u)\\
u \big|_{t = 0} = f \,.
\end{cases}
\end{equation} 
Coefficients $a(t)$ are free parameters (\ie, combination weights) 
to train. $\kappa(u)$ is related to the TV regularization 
\cite{rudin1992nonlinear} and $\mathcal{O}(u)$ denotes a set of 
operators (invariants) over $u$, \eg, $\|\nabla u\|_2^2 = u_x^2 + u_y^2$. 

\subsubsection{Improvements from the side of image statistics/regularization}
\label{improvements}
As shown in~\cite{relations, biased, KrajsekS10, ScharrBH03}, 
there exist a strong connection between anisotropic diffusion models and 
variational models adopting image priors derived from the statistics of natural images. 
Let us consider the discrete version of the PM model \eqref{PM}, where
images are represented as column vectors, \ie, $u \in
\R^{N}$. The discrete PM model is formulated as the following 
discrete PDE with an explicit finite difference scheme
\begin{equation}\label{pmmodel}
  \hspace*{-0.3cm}\frac{u_{t+1} - u_t}{\Delta t} = -\suml{i = \{x,y\}}{}\nabla_i^\top \Lambda (u_t) \nabla_i u_t
  \doteq -\suml{i = \{x,y\}}{}\nabla_i^\top \phi(\nabla_i u_t) \,,
\end{equation}
where matrices $\nabla_x$ and $\nabla_y \in \R^{N \times N}$ are
finite difference approximation of the gradient operators in
$x$-direction and $y$-direction, respectively and $\Delta t$ denotes
the time step. $\Lambda(u_t) \in \R^{N \times N}$ is defined as a
diagonal matrix
\[
\Lambda(u_t) = \text{diag}\left( g\left(\sqrt{(\nabla_x u_t)^2_p +
      (\nabla_y u_t)^2_p}\right)\right)_{p = 1, \cdots, N}\,,
\]
where function $g$ is the edge-stopping function mentioned before. 
If ignoring the coupled relation between $\nabla_x u$
and $\nabla_y u$, the PM model can also be written in the form $\phi(\nabla_i u)
= \left( \phi(\nabla_i u)_1, \cdots, \phi(\nabla_i u)_N \right)^\top
\in \R^N$ with function $\phi(z) = zg(z)$, known as influence function
\cite{ROBUSTANISOTROPIC} or flux function \cite{anisotropicbook}.  
In this paper, we stick to this discrete and decoupled formulation, 
as it is the starting point of our approach. 

As shown in previous works, \eg, \cite{relations, zhu1997prior}, 
the diffusion step \eqref{pmmodel} corresponds to a gradient descent 
step to minimize the energy functional given as 
\begin{equation}\label{energy_pm}
\mathcal{R}(u) = \suml{i \in \{x,y\}}{} \sum\limits_{p = 1}^{N} \rho((k_i * 
u)_p)\,,
\end{equation}
the functions $\rho$ (\eg, $\rho(z) = \text{log}(1+z^2)$) 
is the so-called penalty function.\footnote{Note that $\rho'(z) = \phi(z)$.}  
It is worthwhile to mention that 
the matrix-vector product $\nabla_x u$ can be interpreted as a
2D convolution of $u$ with the linear filter $k_x = [-1, 1]$
($\nabla_y$ corresponds to the linear filter $k_y = [-1, 1]^\top$). 
The energy functional \eqref{energy_pm} can be also understood from the aspects of 
image statistics, image prior and image regularization. 
As a consequence, a lot of efforts listed below 
have been made to improve the capability of model \eqref{energy_pm}.
\begin{itemize}[leftmargin=*]
\setlength\itemsep{0em}
\item[a)] \noindent 
More filters of larger kernel size were considered in \cite{zhu1997prior, 
RothFOE2009, ChenPRB13, Barbu2009}, instead of relatively 
small kernel size, such as usually used 
pair-wise kernels. The resulting regularization model 
leads to the so-called fields of experts (FoE) \cite{RothFOE2009} 
image prior, which works well 
for many image restoration problems. 
\item[b)] Instead of hand-crafted ones with fixed shape, 
more flexible penalty functions were exploited, and they were learned from 
data \cite{zhu1997prior, ScharrBH03, KrajsekS10, CSF2014}. Especially, 
as shown in \cite{zhu1997prior} that those 
unusual penalties such as inverted penalties 
(\ie, $\rho(z)$ decreasing as a function of $|z|$) 
were found to be necessary.
\item[c)] In order accelerate the inference phase related to the model 
\eqref{energy_pm}, in 
\cite{Barbu2009, DomkeAISTATS2012}, it was proposed to truncate the gradient 
descent procedure to fixed iterations/stages, 
and then train this truncated optimization 
algorithm based on the FoE prior model.
\item[d)] In order to further increase the flexibility of multi-stage models, 
Schmidt and Roth \cite{CSF2014} considered varying parameters per stage.
\end{itemize}
\subsection{Motivations and Contributions}\label{motivations}
In this paper we concentrate on nonlinear diffusion process due to its
high efficiency. Taking into consideration the improvements mentioned
in Sec.  \ref{improvements}, we propose a trainable nonlinear
diffusion model with (1) fixed iterations (also referred to as
stages), (2) more filters of larger kernel size, (3) flexible
penalties in arbitrary shapes, (4) varying parameters for each
iteration, \ie, time varying linear filters and penalties. Then all
the parameters (\ie, linear filters and penalties) in the proposed
model are simultaneously trained from training data in a supervised
way. The proposed approach results in a novel learning framework to
train effective image diffusion models. It turns out that the trained
diffusion processes leads to state-of-the-art performance, while
preserve the property of high efficiency of diffusion based
approaches. }
In summary, 
our proposed nonlinear diffusion process offers the following advantages: 
\begin{itemize}
\setlength\itemsep{0em}
    \item[1)] It is conceptually simple as it is merely a standard nonlinear diffusion model with trained filters and influence functions; 
    \item[2)] It has broad applicability to a variety of image restoration problems. 
In principle, all the diffusion based models can be revisited with appropriate training;
    \item[3)] It yields excellent results for several tasks in image restoration, 
including Gaussian image denoising,single image super resolution and JPEG deblocking;
    \item[4)] It is highly computationally efficient, and well suited for parallel computation on GPUs.
\end{itemize}

A shorter paper has been presented as a conference version 
\cite{ChenPock15}. In this paper, 
we incorporate additional contents listed as follows 
\begin{itemize}
\setlength\itemsep{0em}
    \item[1)] We investigate more details of the training phase, such as the influence of (a) initialization, (b) the model capacity and (c) the number of training samples; 
    \item[2)] We consider more detailed analysis of trained models, such as how the trained models generate the patterns;
    \item[3)] We exploit an additional application of single image super resolution 
to further illustrate the potential breadth of our 
proposed learning framework. 
\end{itemize}

\section{Proposed reaction diffusion process}
{In this section, we first describe our learning based reaction 
diffusion model for image restoration, and then 
we show the relations between the proposed model and existing image 
restoration models. 
\subsection{Proposed nonlinear diffusion model}
The fundamental idea of our proposed learning based reaction diffusion 
model is described in Sec. \ref{motivations}. In addition, 
we incorporate a reaction term in order to apply our model for 
different image processing problems, as shown later. 
As a consequence, our proposed nonlinear reaction diffusion model is formulated as
\begin{equation}\label{ourreactionmodel}
\frac{u_{t} - u_{t-1}}{\Delta t} = 
-\underbrace{\suml{i = 1}{N_k}{K_i^t}^\top \phi_i^t(K_i^t u_{t-1})}_\text{diffusion term}~ - ~
\underbrace{\psi^t(u_{t-1}, f)}_\text{reaction term},
\end{equation}
where $K_i \in \R^{N \times N}$ is a highly sparse matrix, implemented as 
2D convolution of the image $u$ 
with the filter kernel $k_i$, i.e., $K_i u \Leftrightarrow k_i*u$, 
$K_i$ is a set of linear filters and $N_k$ is the number of filters. 
In practice, we set $\Delta t = 1$, as we can freely scale the functions 
$\phi_i^t$ and $\psi^t$ on the right hand side. 

Note that our proposed diffusion process is truncated after a few stages, usually 
less than 10. Moreover, the linear filters and influence functions 
are adjustable and vary across stages. As our proposed approach is inspired by nonlinear reaction diffusion model but with trainable filters and 
influence functions, we coin our method Trainable Nonlinear Reaction Diffusion (TNRD). 

In practice, 
we train the proposed nonlinear diffusion model \eqref{ourreactionmodel} 
for specific image restoration problem by exploiting 
application specific reaction terms $\psi(u)$. 
For classical image restoration problems, such as
Gaussian denoising, image deblurring, image super resolution and image
inpainting, we can set the reaction term to be the gradient of a data
term, i.e. $\psi(u) = \nabla_u \cD(u)$. 

For example, if we choose $\cD^t(u, f) =
\frac{\lambda^t}{2} \|A u - f\|_2^2$, we have $\psi^t(u) = \lambda^t A^\top(Au -
f)$, where $f$ is the degraded input image, $A$ is the
associated linear operator, and $\lambda^t$ is related to the strength
of the reaction term. In the case of Gaussian denoising, $A$ is the
identity matrix; for image super resolution, $A$ is related to the down sampling operation and for image 
deconvolution, $A$ corresponds to the linear blur kernel.

\subsection{A more general formulation of the proposed diffusion model}
Note that the data term $\cD(u)$ related to \eqref{ourreactionmodel} 
should be differentiable. In order to handle the problems involving a 
non-differentiable data term, \eg, the JPEG deblocking 
problem investigated in Section \ref{JPEG}, we consider a more general form 
of our proposed diffusion model as follows
\begin{equation}\label{general_form}
\small
u_{t} = \text{Prox}_{\cG^t} \left(
u_{t-1} - \left(\sum\limits_{i = 1}^{N_k}(K_i^t)^\top \phi_i^t(K_i^t u_{t-1}) 
+ \psi^t(u_{t-1}, f)\right)\right),
\end{equation}
where $\text{Prox}_{\cG^t}(\hat u)$ 
is the proximal mapping operation \cite{ipiano} related to 
the non-differentiable function $\cG^t$, given as 
\[
\text{Prox}_{ \cG^t}(\hat u) = \min\limits_{u}\frac{\|u - \hat u\|_2^2}{2} 
+  \cG^t(u) \,.
\]

\vspace*{-0.5cm}
\subsection{Related image restoration models}\label{relations}
\begin{figure*}[htb!]
\centering
\vspace{-0.5cm}
\hspace*{-0.9cm} {\includegraphics[width=1.1\linewidth]{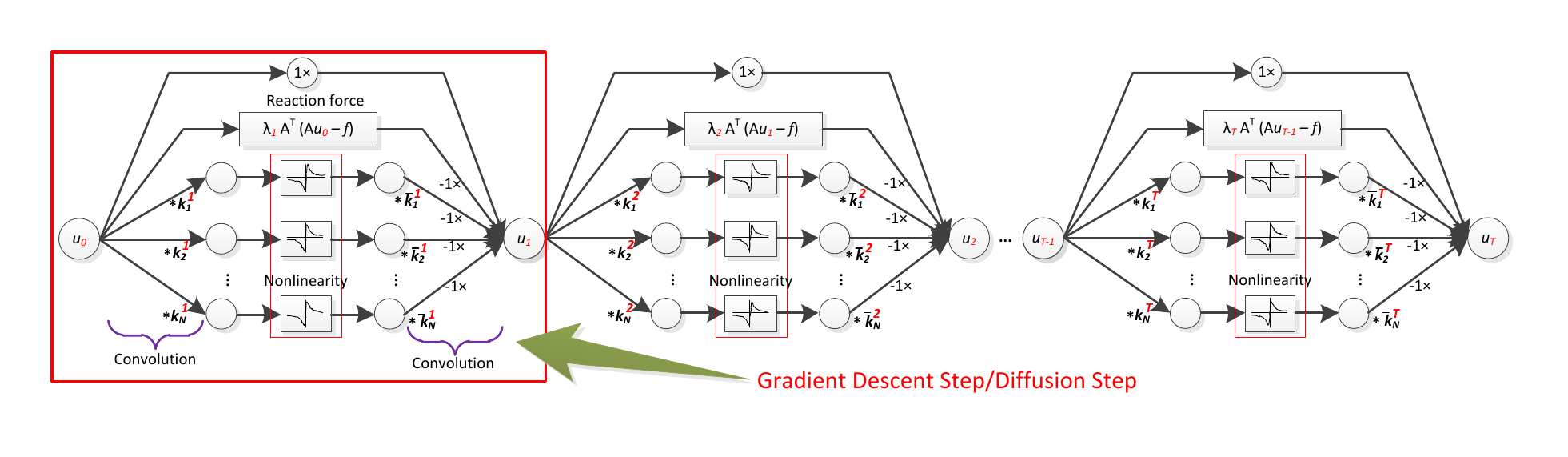}}
\vspace*{-1.2cm}
\caption{The architecture of our proposed diffusion model with a reaction 
force $\psi^t = \lambda^t A^\top (A u - f)$ and $\cG = 0$. 
It is represented as a feed-forward network. Note that the additional convolution step with the rotated kernels $\bar k_i$
(\textit{cf.} Equ. \ref{denoising}) does not appear in conventional feed-forward CNs. }\label{fig:feedforwardCNN}
\end{figure*}
As mentioned in Sec. \ref{improvements},  
there exist natural connection between anisotropic diffusion and 
image regularization based energy functional. Therefore, Eq. \eqref{general_form}
can be interpreted as a forward-backward step\footnote{
The forward step is a gradient descent step w.r.t the function $\cD$ 
and the backward step is the proximal operation w.r.t the function $\cG$.} 
\cite{lions1979splitting} at
$u_{t-1}$ for the energy functional given by
\begin{equation}\label{foemodel}
E^t(u, f) = \suml{i = 1}{N_k} \mathcal{R}^t_i(u) + \cD^t(u, f) + \cG^t(u, f)\,,
\end{equation}
where $\mathcal{R}^t_i(u) = \sum\nolimits_{p = 1}^{N} \rho_i^t((K_i^t
u)_p)$ are the regularizers. 
Since the parameters \{$K_i^t,
\rho_i^t$\} vary across the stages, \eqref{foemodel} is a dynamic energy
functional, which changes at each iteration.

If \{$K_i^t, \rho_i^t$\} keep the same across stages, 
the functional \eqref{foemodel} with $\cG = 0$ corresponds to 
the FoE prior regularized variational model for image
restoration \cite{RothFOE2009, ChenRP14, ChenPRB13}.  In our work, we
do not exactly solve this minimization problem anymore. In
contrast, we run the gradient descent step for several iterations, and
each gradient descent step is optimized by training. 
More importantly, we are allowed to investigate more generalized penalties. 

To the best of our knowledge, Zhu and Mumford \cite{zhu1997prior} were
the first to consider learning reaction diffusion models from data. The linear filters appearing in the prior 
were chosen (not fully trained) from a general filter bank by minimizing the entropy 
of probability distributions of natural images. The associated penalty functions were learned based on the maximum entropy principle. 
However, our proposed diffusion model is a multi-stage diffusion process with 
multiple image priors, where the filters and penalties are fully trained from clean/degraded pairs. 

Some more works have also been dedicated to train the penalties in a diffusion 
model. In \cite{KrajsekS10}, the authors trained the penalty functions in the way that 
they first computed the image statistics appropriate to certain 
chosen filters (\eg, horizontal and vertical image derivatives), and then 
considered mixture models of fixed shape having a peak near zero and two heavy tails 
in both sides to fit to image statistics. 
Analogously, in \cite{ScharrBH03}, potential functions of fixed shape are chosen to 
resemble zero mean filter responses. 

In very recent work \cite{CSF2014}, Schmidt and Roth exploited an additive form of half-quadratic optimization to solve problem \eqref{foemodel}. The resulting model shares similarities with classical wavelet shrinkage and hence it is termed ``cascade of shrinkage fields'' (CSF). 
The CSF model relies on the assumption that the data term in \eqref{foemodel} is quadratic and the operator $A$ can 
be interpreted as a convolution operation, such that the corresponding subproblem can be solved in closed-form using the discrete Fourier transform (DFT). However, our proposed diffusion model does not have this restriction on the data term. In principle, any smooth data term is appropriate. Moreover, as shown in the following sections, we can even handle the case of non-smooth data terms. 

As already mentioned, \cite{Barbu2009, DomkeAISTATS2012} also proposed to train an optimized gradient descent algorithm for the energy functional 
similar to \eqref{foemodel}. However, their model is much more constrained, since they exploited the same 
filters for each gradient descent step and more importantly, they make use of a hand-selected influence function. This clearly restricts the model capability, as demonstrated in Sec. \ref{sec:denoising}. 

Comparing our model to the model of~\cite{liu2010learning}
(cf. \eqref{eq.liu}), one can see that this approach learns only a
linear model based on pre-defined image invariants. Therefore, this
model can be interpreted as a simplified version of our
model~\eqref{ourreactionmodel} with fixed linear filters and influence
functions, and only the weight of each term is optimized.}

\begin{figure}[t!]
\centering
\vspace{-0.5cm}
\hspace*{-0.8cm} {\includegraphics[width=1.15\linewidth]{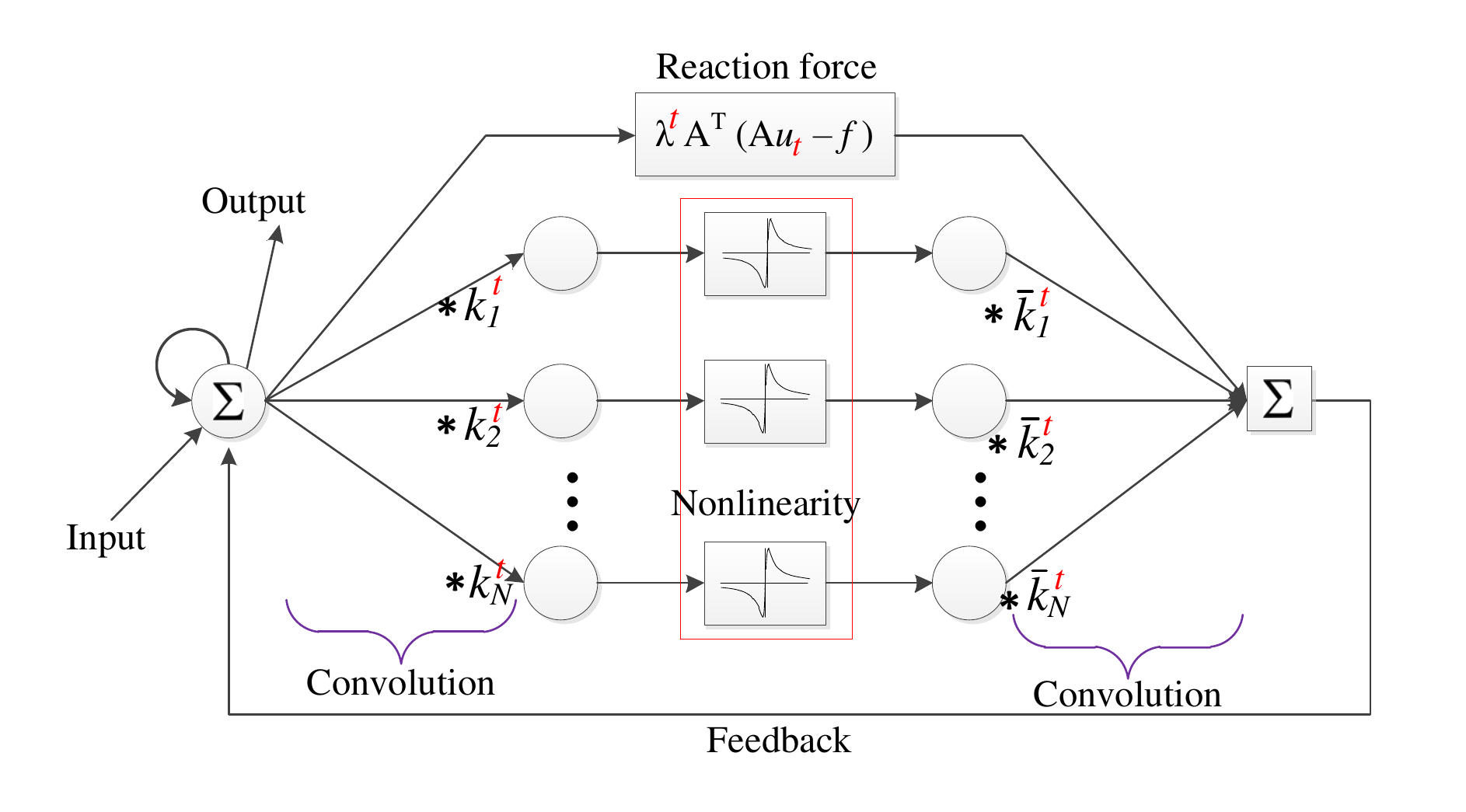}}
\vspace*{-1cm}
\caption{Our diffusion network can also be interpreted as a CN with a 
feedback step, which makes it different from conventional feed-forward networks. Due to the feedback step, 
it can be categorized into recurrent networks \cite{graves2009offline}. 
}\label{fig:recurrentCNN}
\vspace{-0.4cm}
\end{figure}

The proposed diffusion model also bears an interesting link to 
convolutional networks (CNs) applied to 
image restoration problems in \cite{CNNdenoising}. One can see that 
each iteration (stage) of our proposed diffusion process involves convolution 
operations with a set of linear filters, 
and thus it can be treated as a convolutional network. 
The architecture of our proposed diffusion model is shown in Figure \ref{fig:feedforwardCNN}, where it is represented as 
a common feed-forward network. We refer to this network in the following as diffusion network. 

However, we can introduce a feedback step to explicitly illustrate the special 
architecture of our diffusion network that we subtract ``something'' from the input image. Therefore, our diffusion model 
can be represented in a more compact way in Figure \ref{fig:recurrentCNN}, 
where one can see that the structure of our CN model is different from conventional feed-forward networks. 
Due to this feedback step, it can be categorized into recurrent networks \cite{graves2009offline}. 
It should be noted that the nonlinearity (\ie, influence functions in the context of nonlinear diffusion) 
in our proposed network are trainable. However, conventional CNs make use of 
fixed activation function, \eg, the ReLU function 
\cite{nair2010rectified} or sigmoid functions 
\cite{CNNdenoising}. 

\section{Learning framework}
We train our diffusion networks in a supervised manner, namely we firstly prepare the input/output pairs 
for a certain image processing task, and then 
exploit a loss minimization scheme to learn the model parameters $\Theta_t$ for each 
stage $t$ of the diffusion process. The training dataset consists of 
$S$ training samples $\{f^s,u_{gt}^s\}_{s=1}^S$, where 
$f^s$ is a noisy observation and $u_{gt}^s$ is the corresponding ground truth clean image. 
The model parameters $\Theta_t$ of each stage include the parameters of (1) the reaction force weight $\lambda$, 
(2) linear filters and (3) influence functions, \ie, $\Theta_t = \{\lambda^t, \phi_i^t, k_i^t\}$. 

\subsection{Overall training model}
In the supervised manner, a training cost function is required 
to measure the difference between the output of the diffusion 
network and the ground-truth image. As our goal is to train a diffusion network with $T$ stages, the cost function is formulated as 
\begin{equation}\label{joint}
\cL(\Theta_{1, \cdots, T} ) = \suml{s = 1}{S}\ell(u_T^s, u_{gt}^s) \,,
\end{equation}
where $u_T$ is the output of the final stage $T$. In our work we exploit 
the usual quadratic loss function\footnote{
This loss function is related to the PSNR quality measure. Note that as shown in \cite{ECCV2012RTF}, other 
quality measures, such as structural similarity (SSIM) and mean absolute error (MAE) can be chosen to define the loss function.
At present we only consider the quadratic loss function due to its simplicity.}, 
defined as
\begin{equation}\label{loss}
\ell(u_T^s, u_{gt}^s) = \frac 1 2 \|u_T^s - u_{gt}^s\|^2_2 \,.
\end{equation}

\begin{figure*}[t!]
\centering
\includegraphics[width=0.06\textwidth]{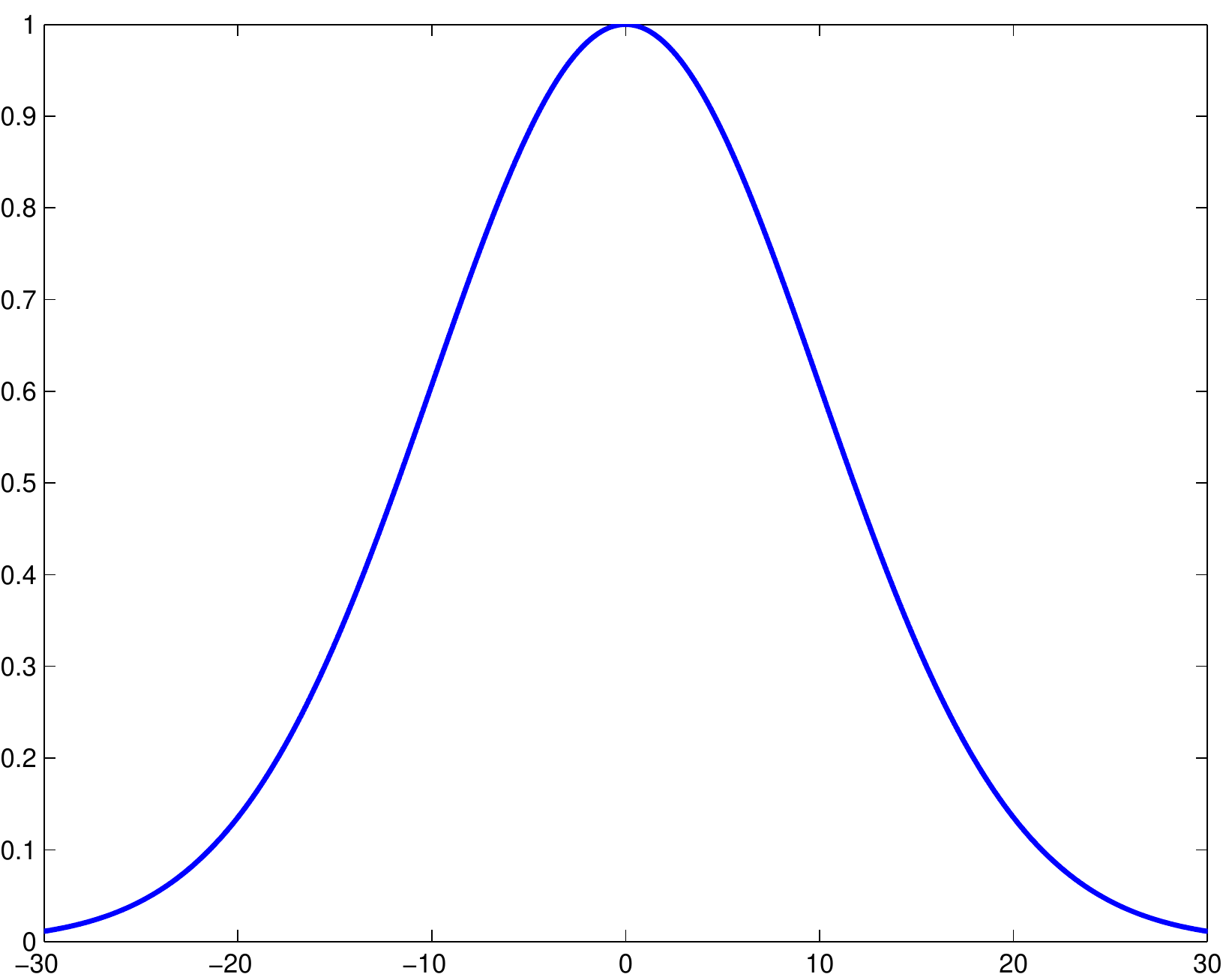}
\includegraphics[width=0.3\textwidth]{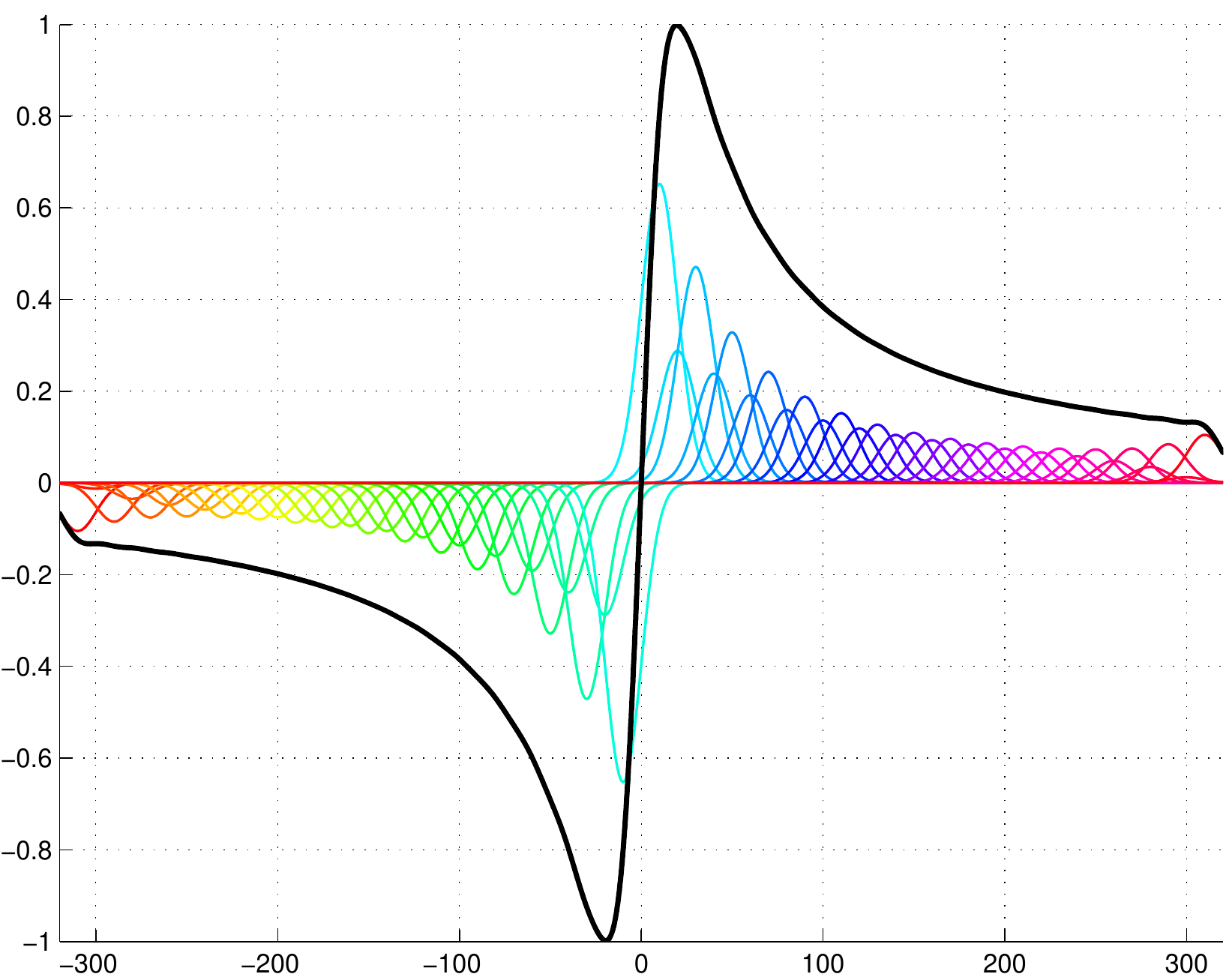}
\includegraphics[width=0.06\textwidth]{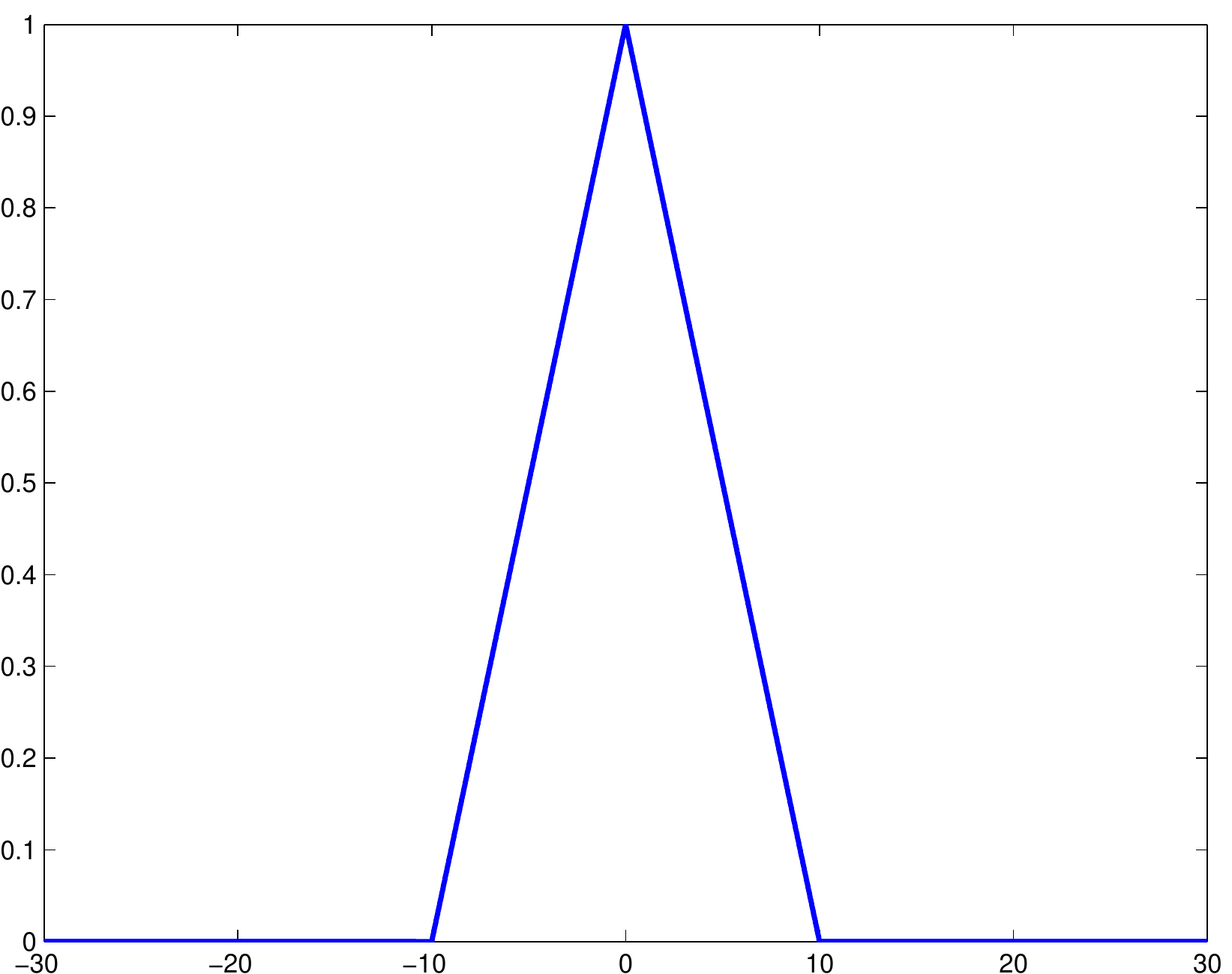}
\includegraphics[width=0.3\textwidth]{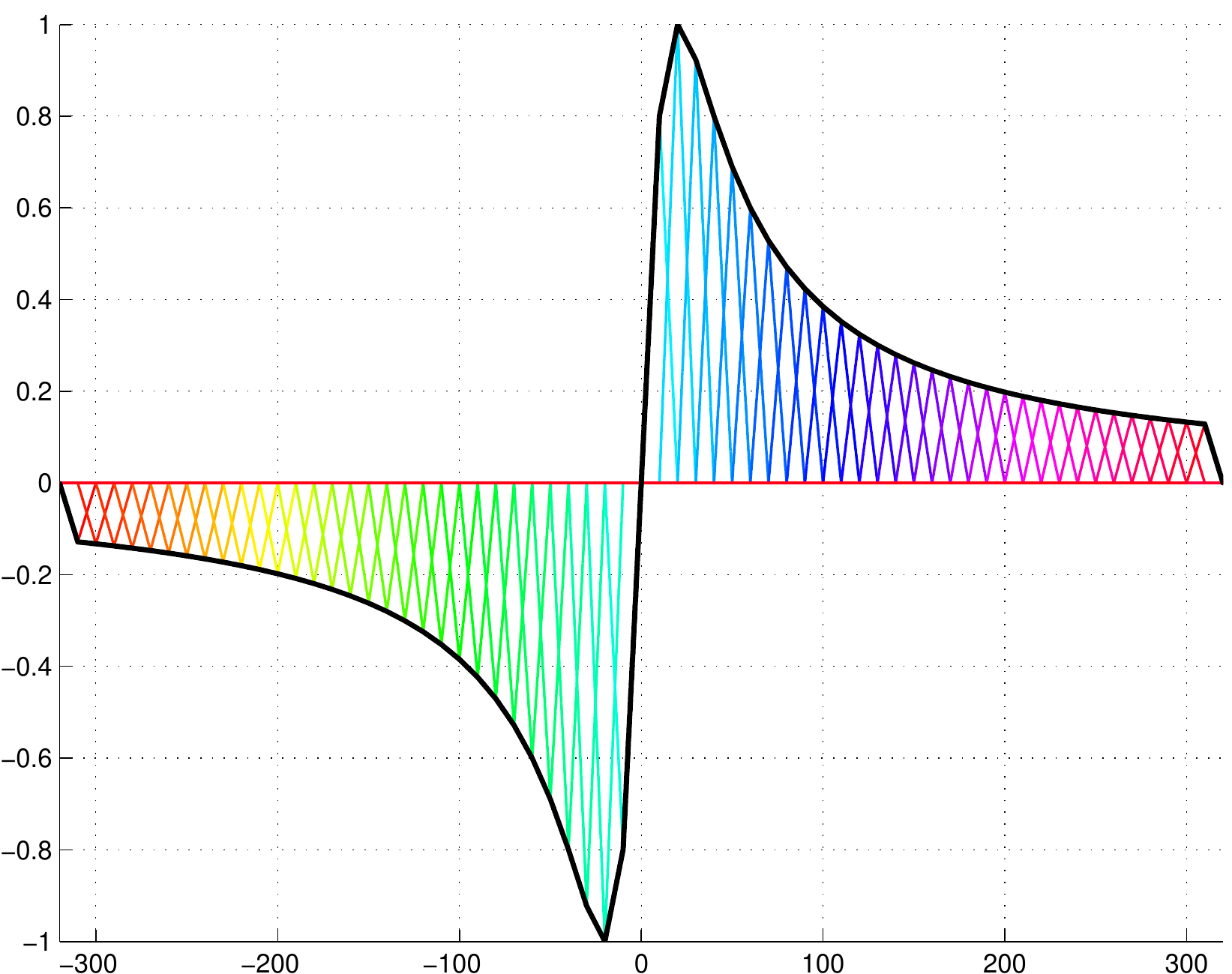}
\caption{Function approximation via Gaussian $\varphi_g(z)$ or 
triangular-shaped $\varphi_t(z)$ radial basis function, respectively for the function $\phi(z) = \frac{2sz}{1+s^2z^2}$ with $s = 
\frac {1}{20}$. Both approximation methods use 63 basis functions equidistantly centered at $[-310:10:310]$.}\label{mapping}
\vspace{-0.4cm}
\end{figure*}

As a consequence, the training task is generally formulated as the following 
optimization problem
\begin{equation}\label{learning}
\footnotesize
\hspace{-0.5cm}\begin{cases}
\min\limits_{\Theta}\cL(\Theta) = \sum\limits_{s = 1}^{S}\frac 1 2
\|u_T^s - u_{gt}^s\|^2_2\\
\text{s.t.}
\begin{cases}
u_0^s = I_0^s \\
u_{t}^s = \text{Prox}_{\cG^t}\left(u_{t-1}^s - 
\left(\sum\limits_{i = 1}^{N_k}(K_i^t)^\top \phi_i^t(K_i^t u_{t-1}^s)
 + \psi^t(u_{t-1}^s, f^s)\right)\right), \\
t = 1 \cdots T\,,
\end{cases}
\end{cases}
\end{equation}
where $\Theta = \{\Theta^t\}_{t=1}^{t=T}$ and $I_0$ is the initial status of 
the diffusion process.
Note that the above loss function only depends on the output of the final 
stage $T$, \ie, the parameters in all stages are 
simultaneously trained such that the output of the diffusion process -  $u_T$ is optimized. {
We call this training scheme \textbf{joint training} similar to \cite{CSF2014}. }
The joint training strategy is a minimization problem with respect to 
the parameters in all stages $\{\Theta_1, \Theta_2, \cdots, \Theta_T\}$.

One can see that our training model is also 
a deep model with many stages (layers). It is well-known that deep models are usually sensitive to initialization, 
and therefore training from scratch is prone to getting stuck at bad local minima. As a consequence, 
people usually consider a greedy layer-wise pre-training \cite{bengio2007greedy} to provide a good 
initialization for the joint training (fine tune). 

{
In our work, we also consider a \textbf{greedy training} scheme similar to 
\cite{CSF2014},} to pre-train our diffusion network stage-by-stage, where 
each stage is greedily trained such that the output of each stage is optimized, \ie, for stage $t$, we minimize the 
cost function
\begin{equation}\label{greedy}
\cL(\Theta_t ) = \suml{s = 1}{S}\ell(u_t^s, u_{gt}^s) \,,
\end{equation}
where $u_t^s$ is the output of stage $t$ of the diffusion process. Note that this is a minimization problem only 
with respect to the parameters $\Theta_t$ in stage $t$. 

\subsection{Parameterizing the influence functions 
$\phi_i^t$, linear filters $k_i^t$ and weights $\lambda^t$}
In this paper, we aim to investigate arbitrary influence functions. 
In order to conduct a fast and accurate training, an effective 
function parameterization method is required. {
Following the work of \cite{CSF2014}, }
we parameterize the influence function via standard radial basis 
functions (RBFs), \ie, each function $\phi$ is represented as a weighted linear combination of a family of RBFs as follows
\begin{equation}\label{rbf}
\phi_i^t(z) = \suml{j = 1}{M}w_{ij}^t\varphi \left(\frac {|z - \mu_j|}{\gamma_j}\right) \,,
\end{equation}
where $\varphi$ represents different RBFs. 
In this paper, we exploit RBFs with equidistant centers $\mu_j$ and unified scaling $\gamma_j$. 
We investigate two typical RBFs \cite{hu2010handbook}: (1) Gaussian radial basis $\varphi_g$ and 
(2) triangular-shaped radial basis $\varphi_t$, given as 
\[
\varphi_g(z) = \varphi \left(\frac {|z - \mu|}{\gamma}\right)  = 
\text{exp}\left(-\frac {(z - \mu)^2}{2\gamma^2}\right)
\]
and 
\[
\varphi_t(z) = \varphi \left(\frac {|z - \mu|}{\gamma}\right)  = 
\begin{cases}
1 - \frac {|z - \mu|}{\gamma}  & |z - \mu| \leq \gamma\\
0 & |z - \mu| > \gamma
\end{cases}
\]
respectively. The basis functions are shown in Figure \ref{mapping}, together with an example of the function approximation 
by using two different RBF methods. 
In our work, we have investigated both function approximation methods, and we find that 
they lead to similar results. 
We only present the results obtained by the Gaussian RBF in this paper. 

In our work, the linear kernels $k_i^t$ related to the linear operators $K_i^t$ are 
defined as a linear combination of Discrete Cosine Transform (DCT) basis kernels 
$b_r$, \ie, 
\[
k_i^t = \frac{\sum\nolimits_{r}\omega_{i,r}^t b_r}{\|\omega_i^t\|_2}\,,
\]
where the kernels $k_i^t$ are normalized to get rid of an ambiguity appearing 
in the proposed diffusion model. 
More details can be found in the \textit{supplemental material}. 
The kernels are formed in this way in order to keep the expected property of 
zero-mean. 

{The weights $\lambda^t$ in our model are constrained to be positive. 
To this end, we set $\lambda^t \leftarrow e^{\lambda^t}$ in the training phase 
for our implementation.} 
\subsection{Computing gradients}
For both greedy training and joint training, we make use of gradient-based algorithms (\eg, 
the L-BFGS algorithm \cite{lbfgs}) for optimization. 
The key point is to compute the gradients of the loss function with respect to the training parameters. 
In greedy training, the gradient of 
the loss function at stage $t$ with respect to the model parameters $\Theta_t$ is computed using standard chain rule, given as
\begin{equation}\label{chainrule}
\frac {\partial \ell(u_t, u_{gt})}{\partial \Theta_t} = 
\frac {\partial u_t}{\partial \Theta_t} \cdot \frac {\partial \ell(u_t, u_{gt})}{\partial u_t} \,,
\end{equation}
where $\frac {\partial \ell(u_t, u_{gt})}{\partial u_t} = u_t - u_{gt}$ is directly derived from \eqref{loss}, 
$\frac {\partial u_t}{\partial \Theta_t}$ is computed from the diffusion process for specific task. For the applications exploited in this 
paper, such as 
image denoising with Gaussian noise, single image super resolution and JPEG deblocking, we present the detailed derivations 
of $\frac {\partial u_t}{\partial \Theta_t}$ in the \textit{supplemental material}.


In the joint training, we compute the gradients of the loss function with respect to $\Theta_t$ by using 
the standard back-propagation technique widely used in the neural networks learning \cite{lecun1998gradient}, namely, 
$\frac {\partial u_t}{\partial \Theta_t}$ is computed by using
\[
\frac {\partial \ell(u_T, u_{gt})}{\partial \Theta_t} = 
\frac {\partial u_t}{\partial \Theta_t} \cdot \frac {\partial u_{t+1}}{\partial u_{t}} \cdots
\frac {\partial \ell(u_T, u_{gt})}{\partial u_T} \,.
\]

Compared to the greedy training, we additionally need to calculate $\frac {\partial u_{t+1}}{\partial u_{t}}$. 
For the investigated image processing problems in this paper, we provide all necessary 
derivations in the \textit{supplemental material}. 

\subsection{Experimental setup and implementation details}
\subsubsection{Boundary condition of the convolution operations}
In our convolution based diffusion network, the image size stays 
the same when an image goes through the network, and 
we use the symmetric boundary condition for convolution calculation. 
In our original diffusion model \eqref{general_form}, there is matrix transpose $K^\top v$, which exactly corresponds to 
the convolution operation $\bar k * v$ ($\bar k$ is obtained by 
rotating the kernel $k$ 180 degrees) in the cases of 
periodic and zero-padding boundary conditions. It should be noted that in the case of symmetric boundary condition used in this paper, 
this result holds only in the central image region. 
However, we still want to explicitly use the formulation $\bar k * v$ to replace $K^\top v$, because the former 
can significantly simplify the derivation of the gradients required for training. 

We find that the direct replacement introduces some artifacts at 
the image boundary. In order to avoid these artifacts, 
we symmetrically pad the input image before it is sent to the diffusion network, and then we discard those padding pixels 
in the final output. More details are found in the \textit{supplemental material}. 

\subsubsection{RBF kernels}
Images exploited in this paper have the dynamic range in $[0, 255]$, and 
the filters have unit norm. 
In order to cover most of the filter response, we consider influence functions in the range $[-310, 310]$. We use 
63 Gaussian RBFs with equidistant centers at $[-310: 10: 310]$, and set the scaling parameter $\gamma = 10$. 

\subsubsection{Experimental setup}
\noindent \textbf{Model capacity}: 
In our work, we train the proposed diffusion network with at most 8 
stages to observe its saturation behavior after certain stages. 
We first greedily train $T$ stages of our model with 
specific model capacity, then conduct a joint training to refine 
the parameters of the whole $T$ stages. 

In this paper, we mainly consider four different diffusion networks with increasing capacity: 
\begin{subequations}
\begin{align*}
        \text{TNRD}_{3 \times 3}^T, ~&\text{Fully trained model with 8 filters of size} ~3 \times 3\,,\\
        \text{TNRD}_{5 \times 5}^T, ~&\text{Fully trained model with 24 filters of size} ~5 \times 5\,,\\
        \text{TNRD}_{7 \times 7}^T, ~&\text{Fully trained model with 48 filters of size} ~7 \times 7\,, \\
  	 \text{TNRD}_{9 \times 9}^T, ~& \text{Fully trained model with 80 filters of size} ~9 \times 9\,,
\end{align*}
\end{subequations}
where $\text{TNRD}_{m \times m}^T$ denotes a nonlinear diffusion process of 
stage $T$ with filters of size $m \times m$. The filters number is $m^2 - 1$, 
if not specified. {For example, $\text{TNRD}_{7 \times 7}^T$ model contains 
$(48\times48 ~(\text{filters}) + 48\times63~ (\text{penalties}) + 1~ (
\lambda))~ \cdot T = 5329\cdot T$ free parameters. }

\noindent \textbf{Training and test dataset}: 
In order to make a fair comparison to previous works, we make use of the same training datasets used in previous works for 
our training, and then evaluate the trained models on commonly used test datasets. For image processing problems 
investigated in this paper, \ie, Gaussian denoising, single image super resolution and JPEG deblocking, we consider the following 
training and test datasets, respectively. 
\begin{itemize}[leftmargin=*]
\setlength\itemsep{0em}
    \item[a)] \noindent Gaussian denoising. Following \cite{CSF2014}, we use the same 400 training images, 
and crop a $180 \times 180$ region from each image, resulting in a total of 400 training samples 
of size $180 \times 180$, \ie, roughly 13 million pixels. We then evaluate the denoising performance of a trained model 
on a standard test dataset of 68 natural images, which is suggested by \cite{RothFOE2009}, and later widely used for Gaussian 
denoising testing. Note that the test images are 
strictly separate from the training datasets. 
    \item[b)] Single image super resolution. 
The publicly available framework of Timofte \etal ~\cite{ANR} provides a perfect base to 
compare single image super resolution algorithms. It includes 91 training images and two test datasets \textbf{Set5} and 
\textbf{Set14}. Many recent state-of-the-art learning based image super resolution approaches \cite{schulter15, SRCNN} 
accomplish their comparison 
based on this framework. Therefore, we also use the same 91 training images. We crop 4-5 
sub-images of size $150 \times 150$ from each training image, according to its size, and this finally gives us 421 training samples. 
We then evaluate the performance of the trained models on the \textbf{Set5} and \textbf{Set14} dataset. 
    \item[c)] JPEG deblocking. We train the diffusion models using the same training images as in the case of Gaussian denoising. 
In the test phase, we follow the test procedure in \cite{ECCV2012RTF} for performance evaluation. 
The test images are converted to gray-value, and scaled by a factor of 0.5, resulting 200 images of size $240 \times 160$. 
\end{itemize}

\subsubsection{Approximate training time}
Note that the calculation of the gradients of the loss function in \eqref{chainrule} 
is very efficient even using 
a simple Matlab implementation, 
since it mainly involves 2D convolutions. The training time 
varies greatly for different configurations. Important factors include (1) model capacity, (2) number of training samples, 
(3) number of iterations taken by the L-BFGS algorithm, 
and (4) number of Gaussian RBF kernels used for function approximation. 
We report the most time consuming cases as follows. 

In training, computing the gradients $\frac{\partial \cL}{\partial \Theta}$ with respect to the parameters of one stage 
for 400 images of size $180 \times 180$ takes about 35s ($\text{TNRD}_{5 \times 5}$), 75s ($\text{TNRD}_{7 \times 7}$) or 
165s ($\text{TNRD}_{9 \times 9}$) using Matlab implementation on a server with CPUs: Intel(R) Xeon E5-2680 @ 2.80GHz 
(eight parallel threads using \textit{parfor} in Matlab, 
63 Gaussian RBF kernels for the influence function parameterization). We typically run 200 L-BFGS 
iterations for optimization. Therefore, the total training time, \eg, for the $\text{TNRD}^5_{7 \times 7}$ model is about 
$5 \times (200 \times 75)/3600 = 20.8h$. Code for learning and inference is available on the authors' homepage 
\url{www.GPU4Vision.org}\footnote{
\url{http://gpu4vision.icg.tugraz.at/binaries/nonlinear-diffusion.zip\#pub94}}. 
For the training of the Gaussian denoising task, we have also accomplished a GPU implementation, 
which is about 4-5 times faster than our CPU implementation. 
\begin{figure*}[t!]
\centering
\subfigure[Noisy $u_0$ (20.17)]{\includegraphics[width=0.16\textwidth]{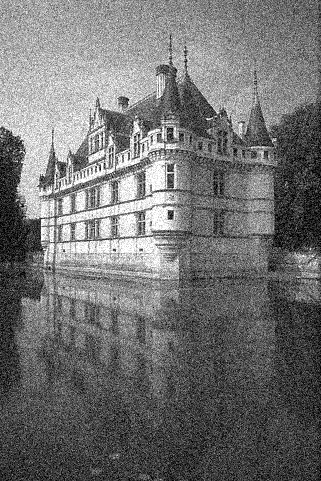}}\hfill
\subfigure[Stage 1: $u_1$ (27.26)]{\includegraphics[width=0.16\textwidth]{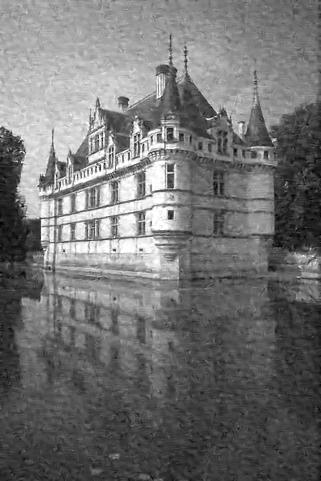}}\hfill
\subfigure[Stage 2: $u_2$ (28.40)]{\includegraphics[width=0.16\textwidth]{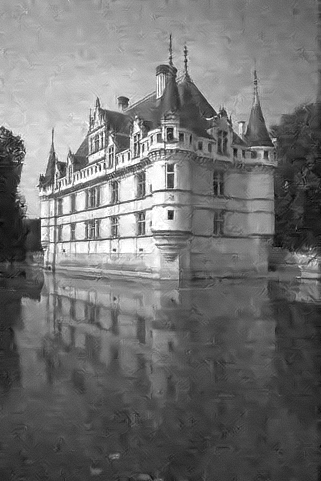}}\hfill
\subfigure[Stage 3: $u_3$ (28.18)]{\includegraphics[width=0.16\textwidth]{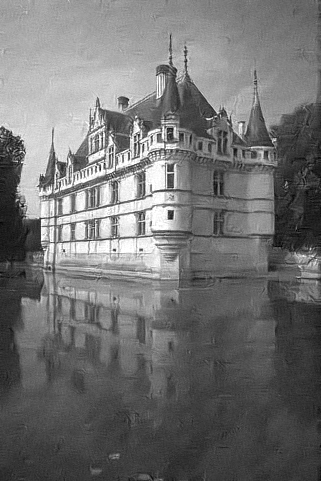}}\hfill
\subfigure[Stage 4: $u_4$ (28.63)]{\includegraphics[width=0.16\textwidth]{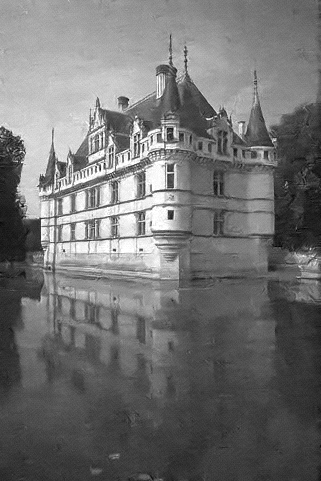}}\hfill
\subfigure[Stage 5: $u_5$ (29.63)]{\includegraphics[width=0.16\textwidth]{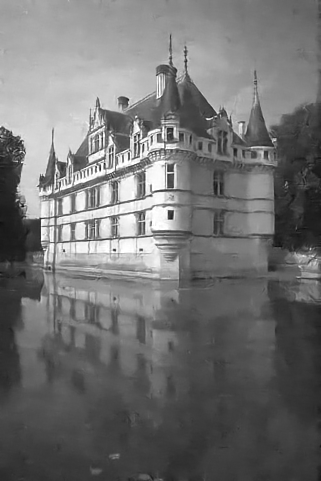}}\\
\vspace*{-0.2cm}
\caption{An image denoising example for noise level $\sigma = 25$ 
to illustrate how our learned $\text{TNRD}_{5 \times 5}^5$ works. (b) - (e) are 
intermediate results at stage 1 - 4, and (f) is the output of stage 5, \ie, the final denoising result.}\label{fig:denoising}
\vspace*{-0.25cm}
\end{figure*}

\section{Training for Gaussian denoising}\label{sec:denoising}
For the task of Gaussian denoising, we consider the following energy functional
\[
\min\limits_{u} E(u) = \suml{i=1}{N_k}\rho_i(k_i * u) + \frac{\lambda}{2}\|u - f\|^2_2\,.
\]
By setting $\cD(u) = \frac{\lambda}{2}\|u - f\|^2_2$ and $\cG(u)  = 0$, 
we arrive at the following 
diffusion process with $u_0 = f$
\begin{equation}\label{denoising}
u_t = u_{t-1} - \left(\sum\limits_{i = 1}^{N_k}\bar k_i^t * \phi_i^t(k_i^t * u_{t-1}) + \lambda^t (u_{t-1} - f)
\right)\,,
\end{equation}
where we explicitly use a convolution kernel $\bar k_i$ (obtained by rotating the kernel $k_i$ 180 degrees) 
to replace the $K_i^\top$ for the sake of model simplicity, but we have to pad the input image. The gradients 
$\frac{\partial u_t}{\partial \Theta_t}$ and $\frac{\partial u_t}{\partial u_{t-1}}$ 
required in training are computed from this equation. Detailed derivations are presented in the \textit{supplemental material}. 

We started with the training for $\text{TNRD}_{5 \times 5}^T$. We first 
considered the greedy training phase to train a 
diffusion process up to 8 stages (\ie, $T \leq 8$), 
in order to observe the asymptotic behavior of the diffusion network. 
{
In the greedy training, only parameters in one stage were trained at a time. 
We exploited a plain initialization to start the training, namely 
linear filters and influence functions were initialized from the 
modified DCT filters and the function $\phi(z) = 2z/(1+z^2)$, respectively. 

After the greedy training was completed, we conducted joint training for a diffusion 
model of certain stages (\eg, $T = 2, 5, 8$)}, 
to simultaneously tune the parameters in all stages.
We initialized the joint training using parameters learned 
in greedy training, as this is guaranteed not to degrade the training performance. 

We first trained our diffusion models for the Gaussian denoising problem with standard deviation $\sigma = 25$. The 
noisy training images were generated by adding synthetic Gaussian noise with $\sigma = 25$ to the clean images. 
Once we obtained a trained model, we evaluated its denoising performance on 68 natural images following the same 
test protocol as in \cite{CSF2014, ChenPRB13}. 
Figure \ref{fig:denoising} shows a denoising example for noise level $\sigma = 25$ 
to illustrate the 
denoising process of our learned $\text{TNRD}_{5 \times 5}^5$ diffusion network.

We present the final results of the joint training 
in Table \ref{denoisingresults}, together with a selection of recent state-of-the-art denoising algorithms, 
namely BM3D \cite{BM3D}, LSSC \cite{LSSC}, EPLL-GMM \cite{EPLL}, opt-MRF \cite{ChenPRB13}, $\text{RTF}$ model 
\cite{ECCV2012RTF}, the CSF model \cite{CSF2014} and WNNM \cite{WNNM}, as well as 
two similar approaches ARF \cite{Barbu2009} and opt-GD \cite{DomkeAISTATS2012}, 
which also train an optimized gradient descent inference. 
We downloaded 
these algorithms from the corresponding author's homepage, and used them as is. 
Unfortunately, we are not able to present comparisons with 
\cite{liu2010learning, CNNdenoising}, as their codes are not available.

From Table \ref{denoisingresults}, one can see that 
(1) the performance of the $\text{TNRD}_{5 \times 5}^T$ model saturates 
after stage 5, \ie, in practice, 5 stages are typically enough; (2) our 
$\text{TNRD}_{5 \times 5}^5$ model has achieved significant improvement (28.78 \vs 28.60), 
compared to a similar model $\text{CSF}_{5 \times 5}^5$, which has the same model capacity and 
(3) moreover, our $\text{TNRD}_{5 \times 5}^8$ model is on par with so far the best-reported algorithm - WNNM. It turns out that our trained models perform 
surprisingly well 
for image denoising. Then, 
a natural question arises: what is the critical factor for the effectiveness 
of the trained diffusion models? 

\subsection{Understanding the proposed diffusion models} 
There are actually two main aspects in 
our training model: (1) the linear filters and (2) the influence functions. In order to have a better understanding of the trained 
models, we went through a series of experiments to investigate the impact of these two aspects. 

Concentrating on the model capacity of 24 filters of size $5 \times 5$, 
we considered the training of a diffusion process with 10 steps, \ie, $T = 10$ for the Gaussian denoising of noise level $\sigma = 25$. 
We exploited two main classes of configurations: 
({\color{blue}{A}}) the parameters of every stage are the same and 
({\color{blue}{B}}) every diffusion stage is different from each other. In both configurations, 
we consider two cases: ({\color{blue}{I}}) only train the linear filters with fixed influence 
function $\phi(z) = 2z/(1+z^2)$ and ({\color{blue}{II}}) simultaneously 
train the filters and influence functions. 

Based on the same training dataset and test dataset, we obtained the following results: 
({\color{blue}{A}}.{\color{blue}{I}}) every diffusion step is the same, and only the filters are optimized with fixed influence 
function. This is a similar configuration to previous works \cite{Barbu2009, DomkeAISTATS2012}. 
The trained model achieves a test performance of 28.47dB. 
({\color{blue}{A}}.{\color{blue}{II}}) with additional tuning of the influence functions, the resulting performance  is boosted to 
28.60dB. ({\color{blue}{B}}.{\color{blue}{I}}) every diffusion step can be different, but only the linear filters are trained with fixed 
influence functions. The corresponding model obtains a result of 28.56dB, which is equivalent to 
the variational model \cite{ChenPRB13} with the same model capacity. Finally ({\color{blue}{B}}.{\color{blue}{II}}) 
with additional optimization of the influence functions, the trained model leads to a significant improvement with the result of 
28.86dB. 

The analytical experiments demonstrate that without the training of the influence functions, there is no chance to achieve 
significant improvements over previous works, 
no matter how hard we tune the linear filters. {Therefore, we believe that 
the most critical factor of our proposed training model lies in the adjustable 
influence functions.} 
A closer look at the learned influence functions of the 
$\text{TNRD}_{5 \times 5}^5$ model in Sec.\ref{influence} strengthens our argument. 
  
Comparing our proposed TNRD model to the CSF model~\cite{CSF2014}, one
can see that the degree of freedom is in principle the same, since in
both models the filters and the non-linear functions can be
learned. Therefore, one would expect a similar performance of both
models in practice. However, it turns out the performance of the CSF
model is inferior to our TNRD model in the case of Gaussian denoising
task. The reason for the performance gap is still unclear and
we plane to investigate it in future work.

\subsection{Learned influence functions} \label{influence}
\begin{figure}[t!]
\vspace*{-0.1cm}\centering
\subfigure[Truncated convex]{\includegraphics[width=0.235\textwidth]{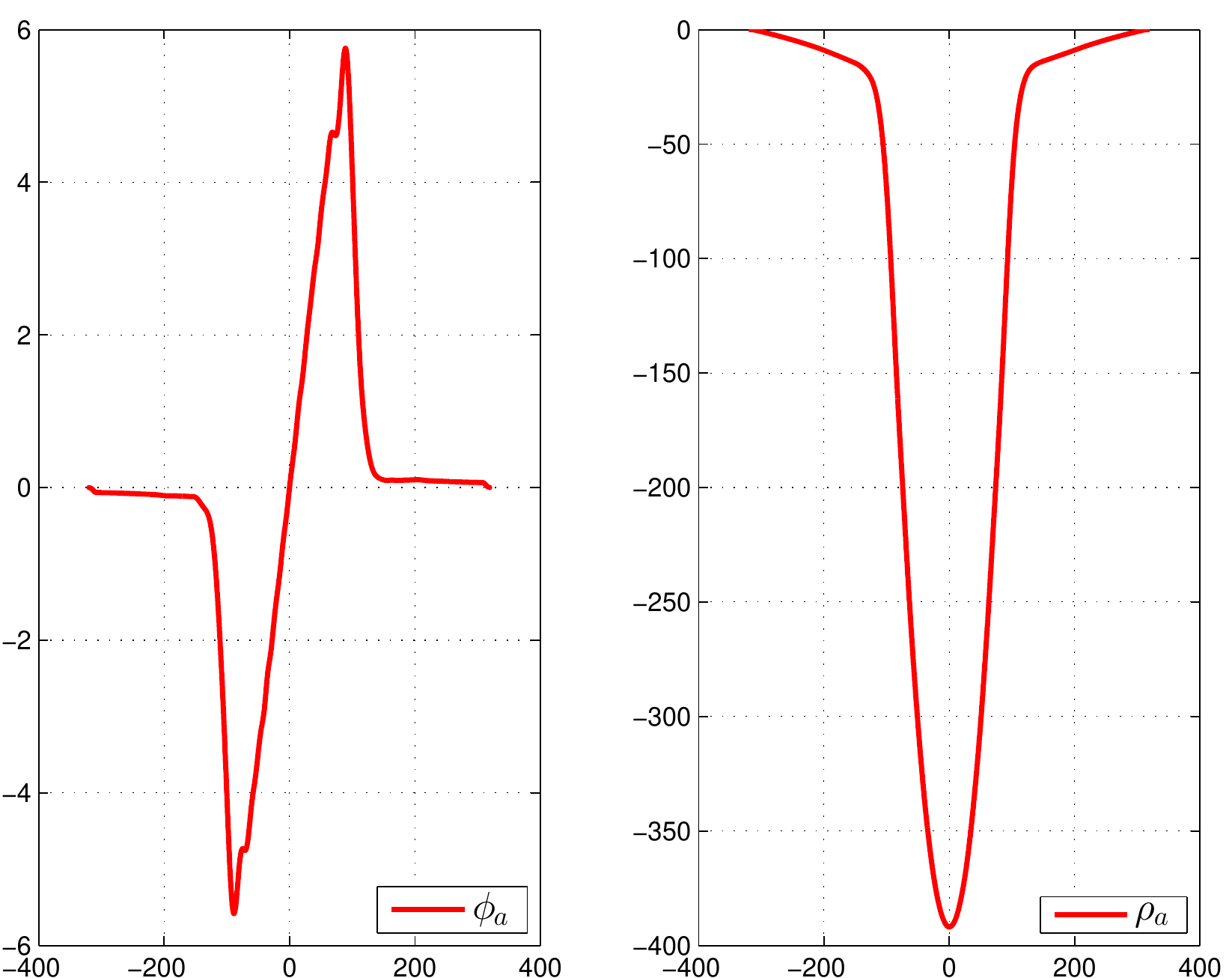}}\hfill
\subfigure[Negative Mexican hat]{\includegraphics[width=0.235\textwidth]{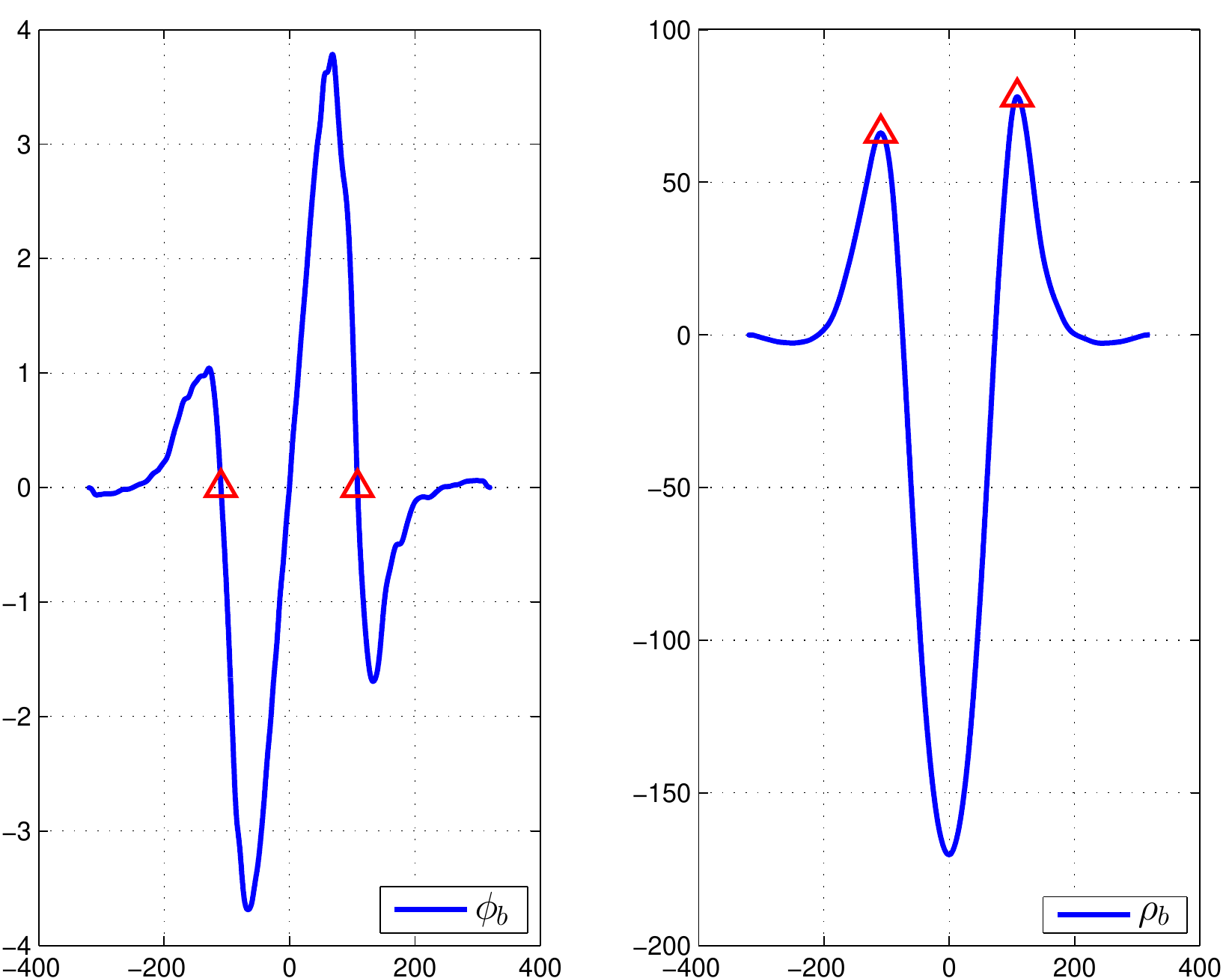}}\\
\vspace*{-0.3cm}
\subfigure[Truncated concave]{\includegraphics[width=0.235\textwidth]{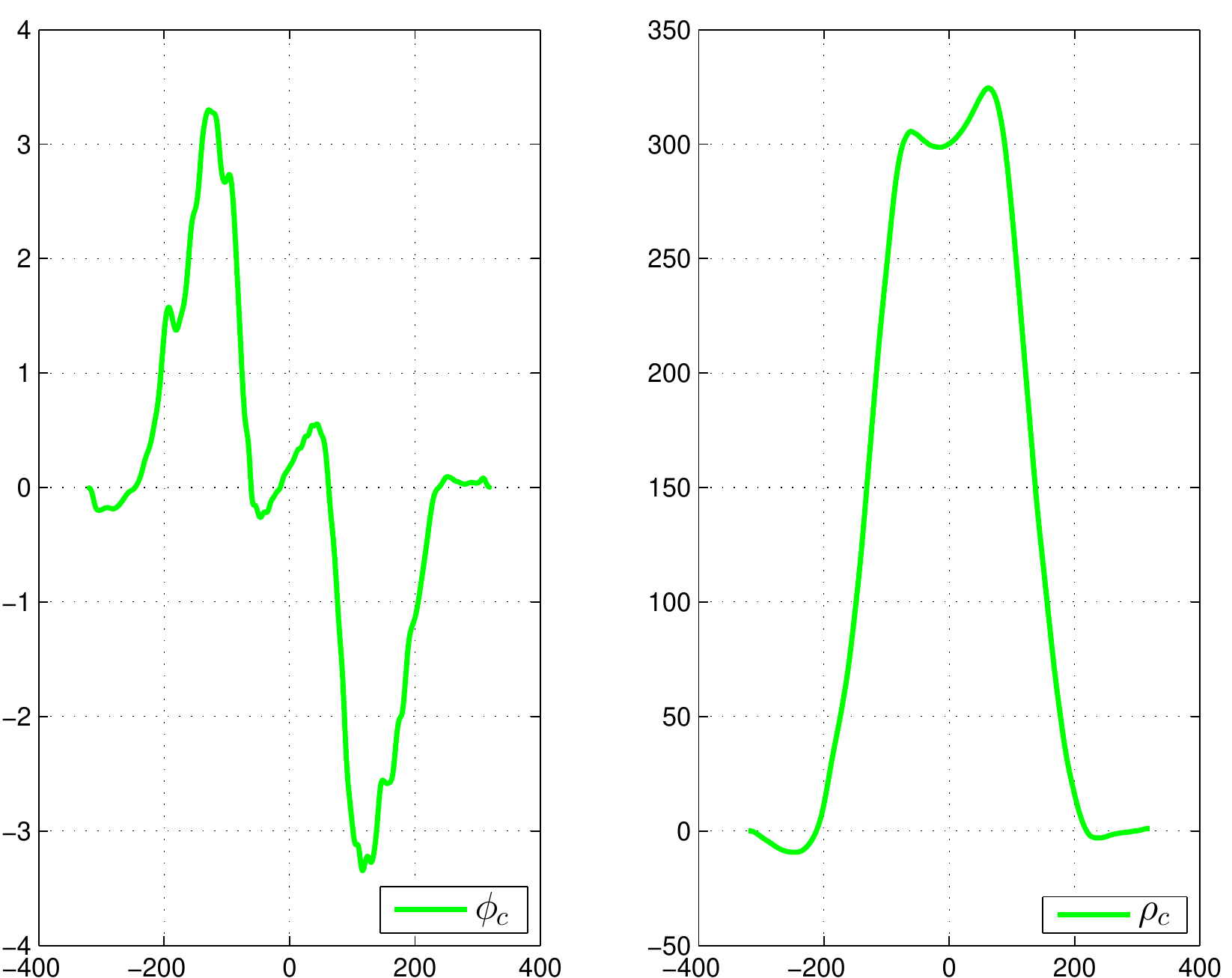}}\hfill
\subfigure[Double-well penalty]{\includegraphics[width=0.235\textwidth]{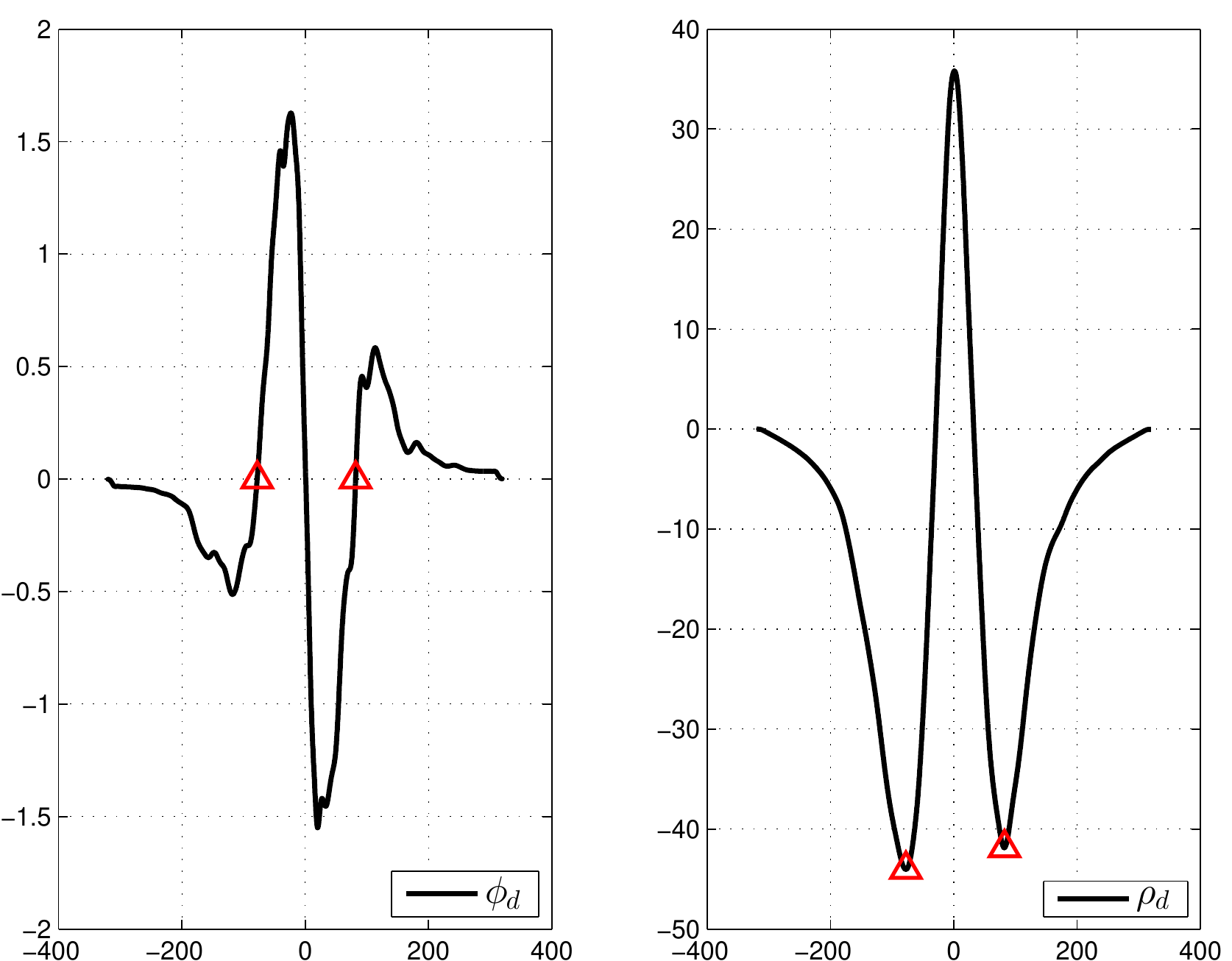}}
\caption{The figure shows four characteristic influence functions
  (left plot in each subfigure) together with their corresponding
  penalty functions (right plot in each subfigure), learned by our
  proposed method in the $\text{TNRD}_{5 \times 5}^5$ model. A major finding in this paper is that our learned
  penalty functions significantly differ from the usual penalty
  functions adopted in partial differential equations and energy minimization
  methods. In contrast to their usual robust smoothing properties
  which is caused by a single minimum around zero, most of our learned
  functions have multiple minima different from zero and hence are
  able to enhance certain image structures. See Sec. \ref{sec:patterns} for more
  information.}\label{functions}
\vspace{-0.25cm}
\end{figure}
{
A close inspection of the learned 120 penalty functions\footnote{The penalty function $\rho(z)$ is integrated from the influence function $\phi(z)$ according 
to the relation $\phi(z) = \rho'(z)$} $\rho$ in the $\text{TNRD}_{5 \times 5}^5$
model demonstrated that most of the penalties resemble four representative shapes 
shown in Figure~\ref{functions}.}
\begin{itemize}
\setlength\itemsep{0em}
\item[(a)] Truncated convex penalty functions with low values around
  zero to promote smoothness.
\item[(b)] Negative Mexican hat functions, which have a local minimum
  at zero and two symmetric local maxima.
\item[(c)] Truncated concave functions with smaller values at the two
  tails.
\item[(d)] Double-well functions, which have a local maximum (not a
  minimum any more) at zero and two symmetric local minima.
\end{itemize}
At first glance, the learned penalty functions (except (a)) differ
significantly from the usually adopted penalty functions used in PDE
and energy minimization methods. However, it turns out that they have
a clear meaning for image regularization.

Regarding the penalty function (b), there are two critical points (indicated by red triangles). When the magnitude of the 
filter response is relatively small (\ie, less than the critical points), 
probably it is stimulated by the noise and therefore the penalty function encourages 
smoothing operation as it has a local minimum at zero. However, once the magnitude of the filter response is large enough 
(\ie, across the critical points), the corresponding local patch probably contains a real image edge or certain structure. 
In this case, the penalty function encourages to increase the magnitude of the filter response, alluding to an image sharpening 
operation. Therefore, the diffusion process controlled by the influence function (b), can adaptively switch between 
image smoothing (forward diffusion) and sharpening (backward diffusion). We find that the learned influence function (b) is closely 
similar to an elaborately designed function in a previous work \cite{gilboa2002forward}, 
which leads to an adaptive forward-and-backward diffusion process. 

Similar forms of the learned penalty function in (c) with a concave shape 
are also observed in previous work on image prior learning \cite{zhu1997prior}. This penalty function also encourages to 
sharpen the image edges. Concerning the learned penalty function (d), as it has local minima at two specific points, it 
prefers specific image structure, implying that it helps to form certain image structure. We also find that this penalty function is 
exactly the type of bimodal expert functions for texture synthesis employed in \cite{HeessWH09}. 

{Therefore, our learned penalty functions confirmed existing robust penalties based 
prior models and many priors exploiting some unusual penalties, which  
can produce patterns and enhance preferred features.} 
As a consequence, the diffusion process involving the learned 
influence functions does not perform 
pure image smoothing any more for image processing. 
In contrast, it leads to a diffusion process for adaptive image smoothing 
and sharpening, distinguishing itself from previous commonly used image regularization techniques. 

\begin{figure*}[t!]
\centering
\hspace*{-1.5cm} {\includegraphics[width=0.8\linewidth]{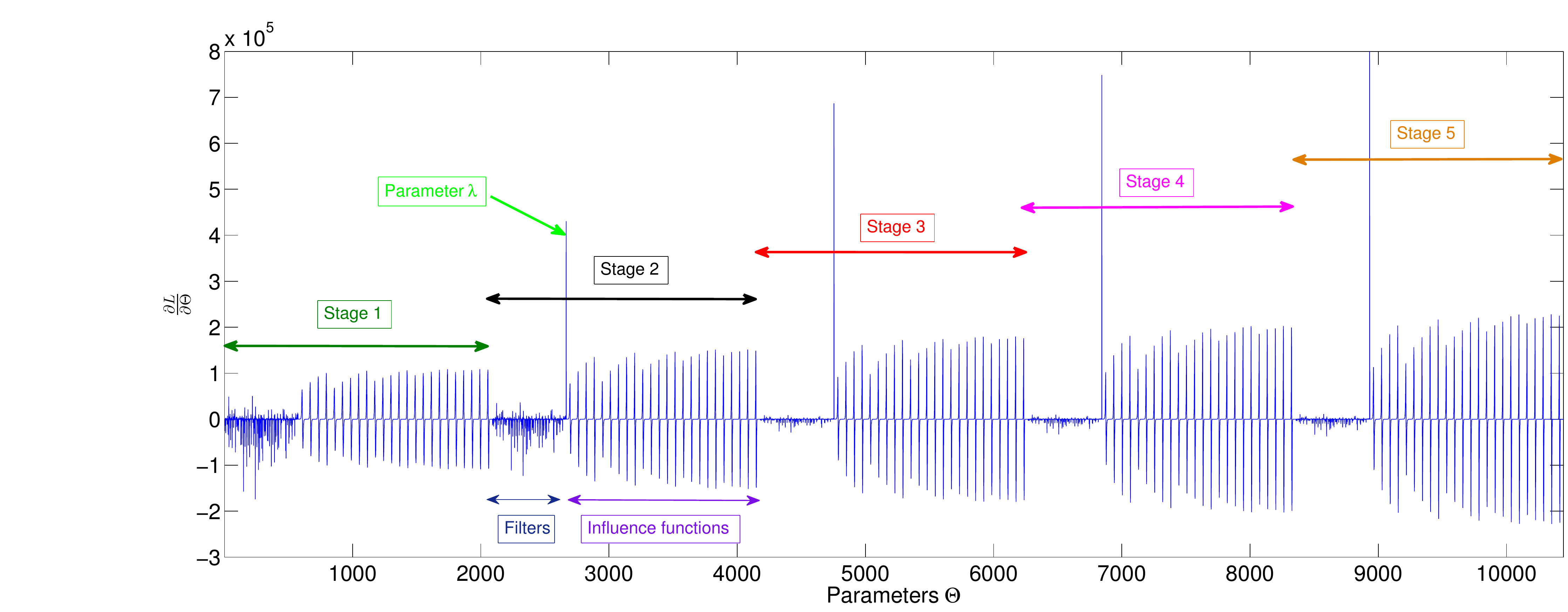}}
\vspace*{-0.25cm}
\caption{Well distributed gradients $\frac{\partial L}{\partial \Theta}$ 
over stages for the $\text{TNRD}_{5 \times 5}^5$ model at the initialization point $\Theta_0$ with a plain setting. 
{One can see that the ``vanishing gradient'' phenomenon 
\cite{bengio1994learning} in the back-propagation phase 
of a conventional deep model does not appear in our training 
model.}}\label{fig:gradients}
\end{figure*}
\subsection{Pattern formation using the learned influence 
functions}\label{sec:patterns}
\begin{figure}[t!]
\centering
\subfigure[]{\includegraphics[width=0.32\linewidth]{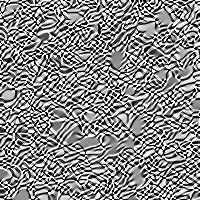}}\hfill
\subfigure[]{\includegraphics[width=0.32\linewidth]{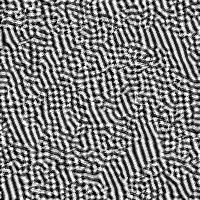}}\hfill
\subfigure[]{\includegraphics[width=0.32\linewidth]{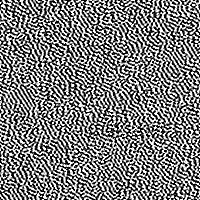}}
\vspace*{-0.25cm}
\caption{Patterns synthesized from uniform noise 
using our learned diffusion models. (a) is generated by \eqref{diffusion} using 
the parameters (linear filters and influence functions) in a stage of our learned $\text{TNRD}_{5 \times 5}^5$ 
for image denoising, (b) is generated by \eqref{diffusion} using 
the parameters in a stage of our learned $\text{TNRD}_{7 \times 7}^5$ for image super resolution and 
(c) is also from a stage of our learned $\text{TNRD}_{7 \times 7}^5$ for image super resolution.}\label{patterns}
\vspace*{-0.5cm}
\end{figure}

In the previous work on Gibbs reaction diffusion\footnote{The terminology of ``reaction diffusion'' in \cite{zhu1997prior} is a bit 
different from ours. In our formulation, ``reaction term'' is related to the data term, while in \cite{zhu1997prior}, it means 
the diffusion term controlled by those downright penalty functions.} 
\cite{zhu1997prior}, it is shown that those unconventional penalty functions 
such as Figure \ref{functions}(c) have significant meaning in visual computation, as they can produce patterns. 
We also find that those unconventional penalty functions learned in our models 
can produce some interesting image patterns. 

We consider the following diffusion process involving our learned linear filters and the corresponding influence functions
\begin{equation}\label{diffusion}
\frac{u_{t} - u_{t-1}}{\Delta t} = 
-\sum\limits_{i = 1}^{N_k}\bar k_i * \phi_i(k_i * u_{t-1}) \,,
\end{equation}
where the filters $k_i$ and influence functions $\phi_i$ are chosen from a certain stage of the learned models. Note that we 
do not incorporate a reaction term in this diffusion model. We run \eqref{diffusion} 
from starting points $u_0$ (uniform 
noise images in the range $[0, 255]$), and it converges to a local minimum\footnote{The corresponding diffusion processes are 
unstable, and therefore we have to restrict the image dynamic range to $[0, 255]$.}. Some synthesized patterns are 
shown in Figure \ref{patterns}. One can see that the diffusion model with our learned influence functions and filters can 
produce edge-like image structure and repeated patterns from fully random images. 
This kind of diffusion model is known as Gibbs reaction diffusion in \cite{zhu1997prior}. 
We provide another example in Figure \ref{fig:SRtoy} 
to demonstrate how our learned diffusion models can generate meaningful patterns for image super resolution. 

\subsection{Important aspects of the training framework}\label{important}
\subsubsection{Influence of initialization}
Our training model is also a deep model with many stages (layers), but we find that it 
is not very sensitive to initialization. 
{Based on the training for Gaussian denoising, 
we conducted experiments with fully random initializations and some plain settings. 

We firstly investigated the case of greedy training where the model was trained 
stage by stage and the parameters of one stage 
were trained at a time. We initialized the parameters using fully random numbers 
in the range $[-0.5, 0.5]$. It turns out that the resulting models with different 
initializations lead to a deviation within 0.01dB in the test phase. That is to say, 
the greedy training strategy is not sensitive to initialization. 

Then, we considered the case of joint training, where all the parameters 
in all stages were trained at a time. 
We also initialized the training with 
fully random numbers in the range $[-0.5, 0.5]$. 
In this case, it turns out that the resulting 
models lead to inferior results, 
e.g., in the case of $\text{TNRD}^5_{5\times 5}$ (28.61 \vs 28.78). 
However, plain initializations can generate equivalent results. For example, 
we considered a plain initialization (all stages were initialized from the modified 
DCT filters and an unified influence function $\phi(z) = 2z/(1+z^2)$), the 
resulting models performed almost the same as those models trained from some 
good initializations such as parameters obtained from greedy training, 
e.g.,$\text{TNRD}^5_{5\times 5}$, (28.75 \vs 28.78) and 
$\text{TNRD}^5_{7\times 7}$, (28.91 \vs 28.92).} 

We believe that this appealing property of our training framework is attributed to the well-distributed gradients across stages. 
We show in Figure \ref{fig:gradients} an example to illustrate 
the gradients of the training loss function with respect to the parameter of all stages. 
One can see that the well-known phenomenon of ``vanishing gradient'' \cite{bengio1994learning} in the back-propagation phase 
of a usual deep model does not appear in our training model. We believe that the 
reason for the well-distributed gradients is that our training model is 
more constrained. For example, in a more general sense, the rotated kernel $\bar k_i$ in our formulation is not necessary to be 
the rotated version of the kernel $k_i$, and it can be an arbitrary kernel. However, we stick to this form, as it has a clear meaning 
derived from energy minimization. 
\subsubsection{Influence of the number of training samples}
\begin{figure}[t!]
\centering
\vspace*{-0.25cm}
   \hspace*{-0.6cm} \includegraphics[width=0.75\linewidth]
{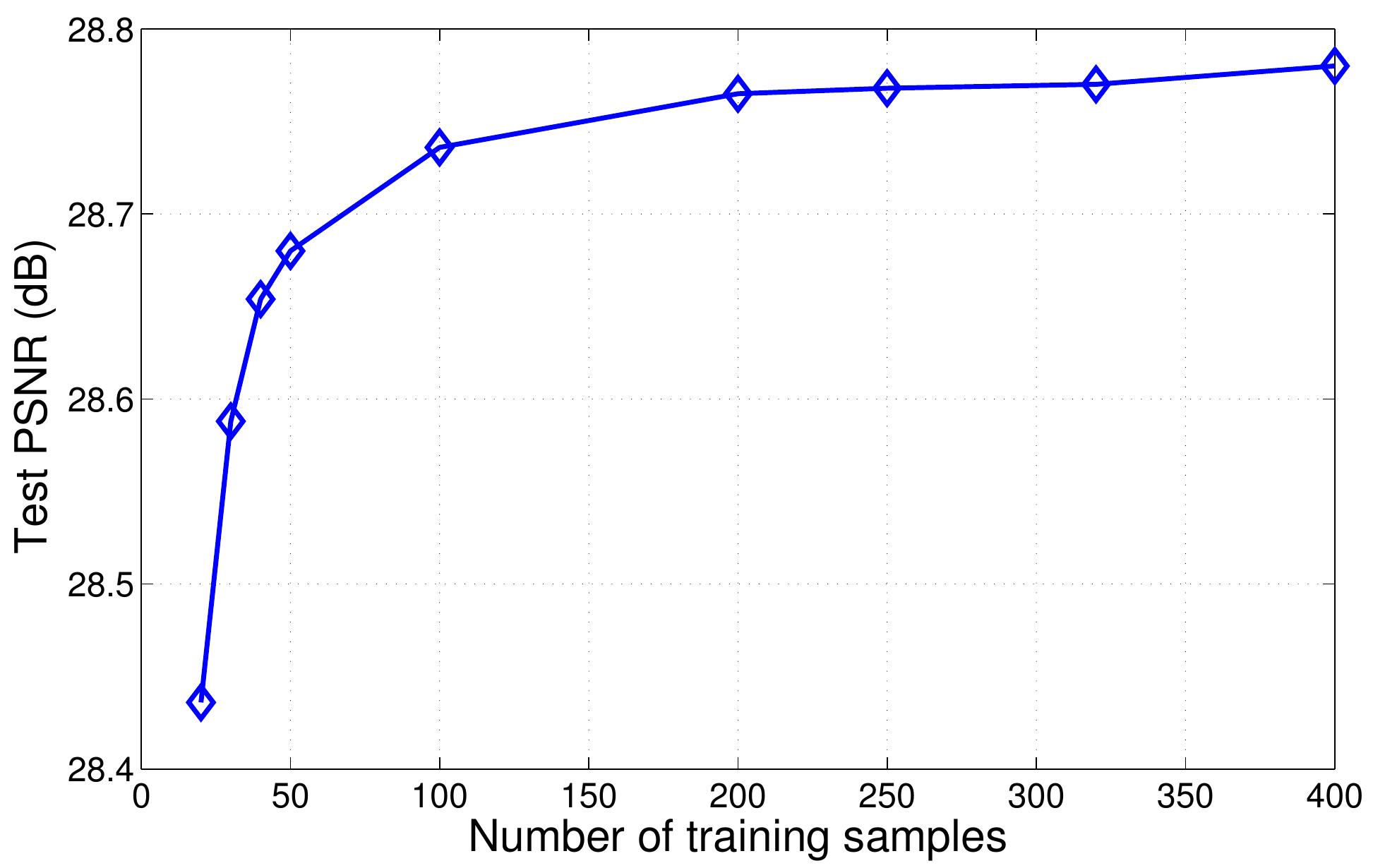}
\vspace*{-0.25cm}
    \caption{{Influence of the number of training examples 
for the training model 
$\text{TNRD}^5_{5\times 5}$}}
\label{fig:samples}
\end{figure}
In our training, we do not consider any regularization for the training 
parameters, and we finally reach good-performing 
models. A probable reason is that we have exploited sufficient training samples (400 samples of size $180 \times 180$). Thus 
an interesting question arises: how many samples are sufficient for our training? 

In order to answer this question, we re-train the $\text{TNRD}^5_{5\times 5}$ model using different size of training dataset, and 
then evaluate the test performance of trained models. We summarize the results in Figure \ref{fig:samples}. 
One can see that (1) 
too few training samples (\eg, 40 images) will clearly lead to over-fitting, thus inferior test performance, and (2) 
200 images are typically enough to prevent over-fitting. 

\subsubsection{Influence of the filter size}
\begin{figure}[t!]
\centering
   \hspace*{-0.6cm} \includegraphics[width=0.75\linewidth]{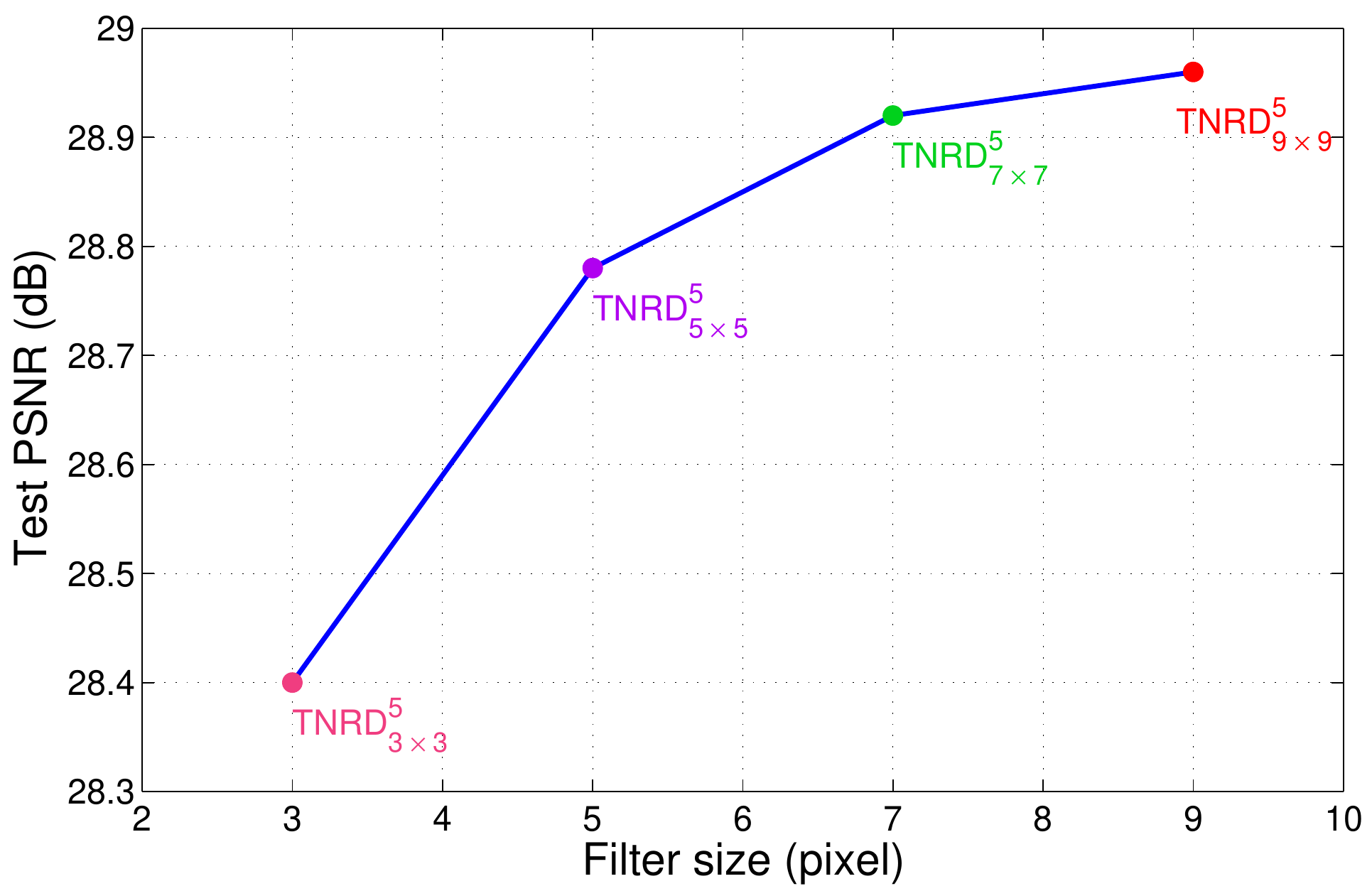}
\vspace*{-0.1cm}
    \caption{{Influence of the filter size 
(based on a relatively small training data set of 400 images of 
size $180 \times 180$)}}
\label{fig:filtersize}
\end{figure}

\begin{figure}[t!]
\centering
\subfigure[48 filters of size $7 \times 7$ in stage 1]{\includegraphics[width=1\linewidth]{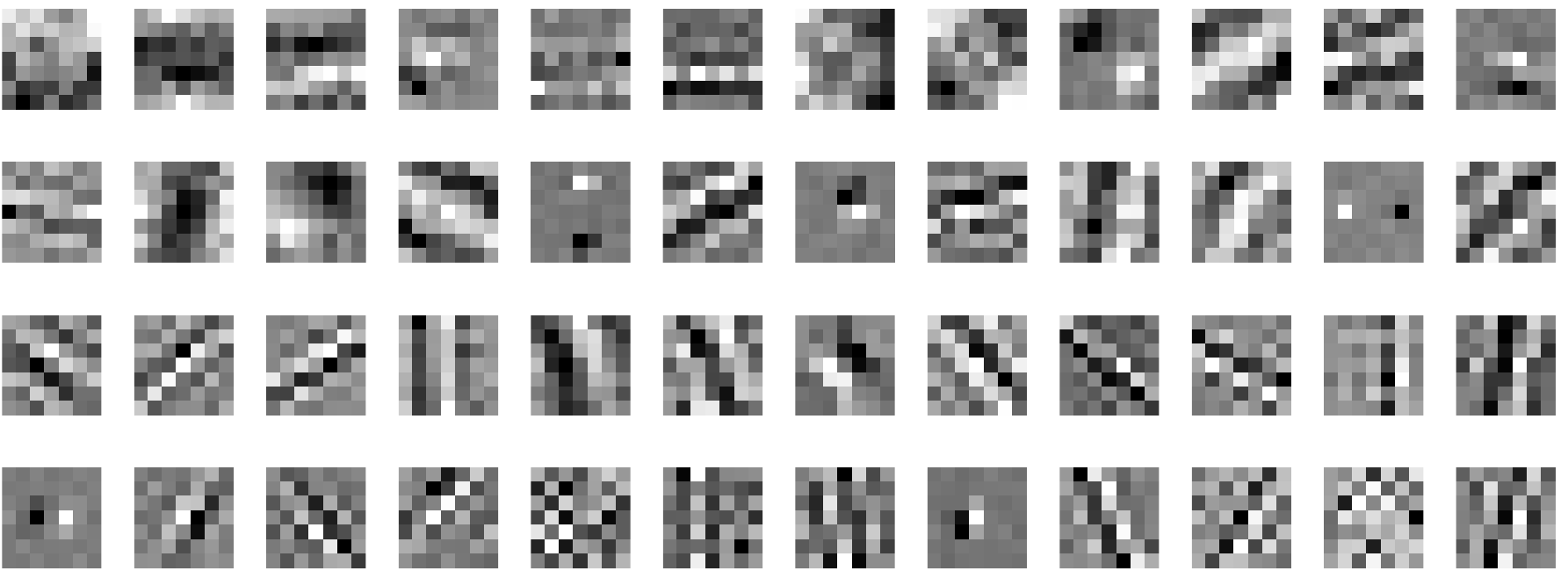}}\\
\vspace*{-0.25cm}
\subfigure[48 filters of size $7 \times 7$ in stage 5]{\includegraphics[width=1\linewidth]{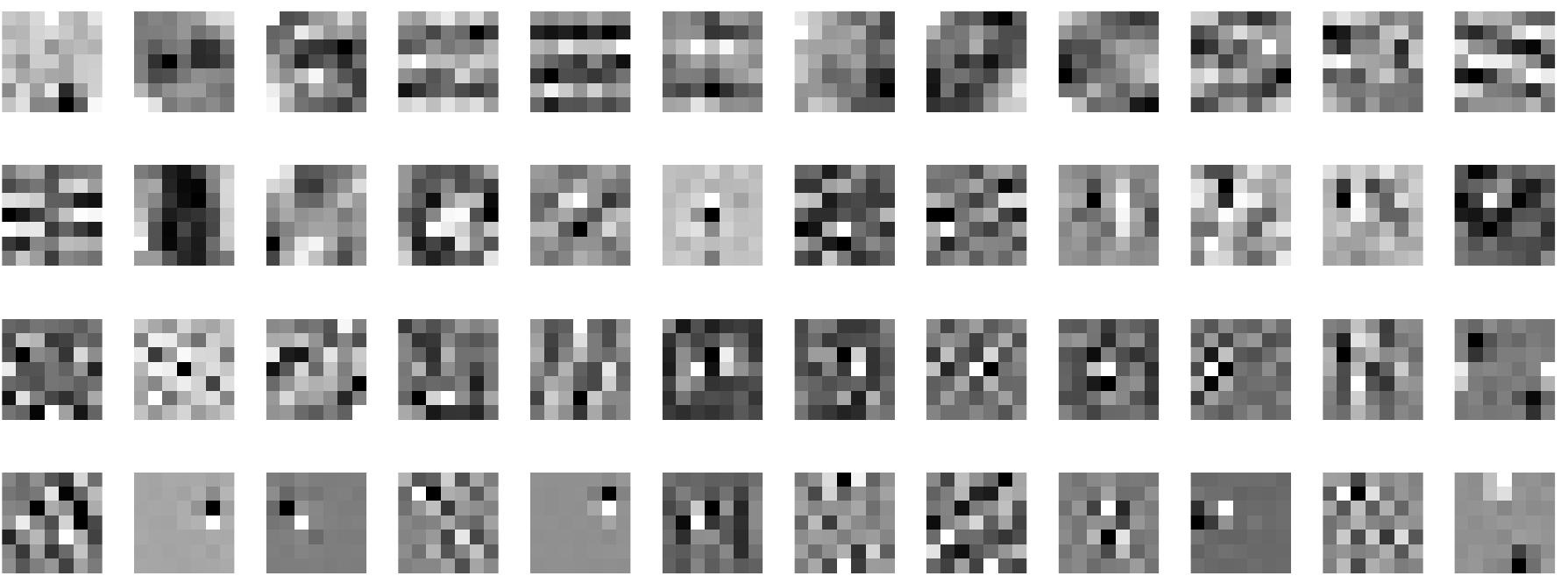}}
\vspace*{-0.25cm}
\caption{Trained filters (in the first and last stage) of the $\text{TNRD}^5_{7\times 7}$ model for the noise level $\sigma = 25$. 
We can find first, second and higher-order derivative filters, as well as rotated derivative filters along different directions. These filters 
are effective for image structure detection, such as image edge and texture.}\label{fig:filters}
\vspace*{-0.25cm}
\end{figure}
In our model, the size of involved filters is a free parameter. 
In principle, we can exploit filters of any size, but in practice, we 
need to consider the trade-off between run time and accuracy. 

In order to investigate the influence of the filter size, we increase the filter size to $7 \times 7$ and $9 \times 9$. We find 
that increasing the filter size from $5 \times 5$ to $7 \times 7$ brings a significant improvement of 0.14dB (
$\text{TNRD}^5_{7\times 7}$ \vs $\text{TNRD}^5_{5\times 5}$) as show in Table \ref{denoisingresults}. 
However, when we further increase the filter size to $9 \times 9$, the resulting $\text{TNRD}^5_{9\times 9}$ only leads to a 
performance of 28.96dB (a slight improvement of 0.05dB relative to the $\text{TNRD}^5_{7\times 7}$ model). 
We can conjecture that further increasing the filter size to $11 \times 11$ 
might bring negligible improvements. 

{Note that the above conclusion is drawn from a relatively small training 
data set of 400 images of size $180 \times 180$. It should be 
mentioned that when the size of the model increase, the size of training data set 
should also increase to avoid over-fitting. 
However, our current CPU implementation for training 
prevents us from training with larger model and large-scale data sets (millions). 
A faster implementation on GPUs together with the stochastic gradient descent 
optimization strategy is left to future work.}

We also consider a model with 
smaller filters, $3 \times 3$. We summarize the results of different model capacities in Figure \ref{fig:filtersize}. 
In practice, we prefer the $\text{TNRD}^5_{7\times 7}$ 
model as it provides the best trade-off between performance and computation time. 
Therefore, in later applications, we only consider $\text{TNRD}^T_{7\times 7}$ models. 

Fig. \ref{fig:filters} shows the trained filters 
of the $\text{TNRD}^5_{7\times 7}$ model in the 
first and last stage for the task of Gaussian denoising. One can find many edge and image structure detection filters along 
different directions and in different scales. 

\subsection{Training for different 
noise levels and comparison to recent state-of-the-arts}
The above training experiments are based on Gaussian noise of level $\sigma = 25$. 
We also trained diffusion models for the noise levels $\sigma = 15$ and $\sigma = 50$. 
The test performance is summarized in Table \ref{denoisingresults}, together with comparison to very recent 
state-of-the-art denoising algorithms. 
In experiments, we observed that joint training can always gain an improvement of about 0.1dB 
over the greedy training for the cases of $T \geq 5$. 

From Table \ref{denoisingresults}, one can see that for all noise levels, 
the resulting $\text{TNRD}_{7\times 7}$ model achieves the highest average PSNR. 
The $\text{TNRD}^5_{7\times 7}$ model outperforms the benchmark - 
BM3D method by 0.35dB in average. This is a notable improvement as few methods can surpass BM3D more 
than 0.3dB in average \cite{levin2011natural}. Moreover, the $\text{TNRD}^5_{7\times 7}$ model also surpasses 
the best-reported algorithm - WNNM method, which is quite slow as shown in Table \ref{run time}. 

\begin{table}[t!]
\centering
\vspace*{0.2cm}
\hspace*{-0.3cm}
\begin{tabular}{l c c c c l l}
\Xhline{0.5pt}
\hline
\multirow{2}*{Method} & \multicolumn{3}{c}{$\sigma$} &\multirow{2}*{\color{blue}St.} 
& $\text{TNRD}_{5 \times 5}$ & $\text{TNRD}_{7\times 7}$\\
\cline{2-4}
\cline{6-7}
&15 & 25 &50 & &\multicolumn{2}{c}{${\sigma = 15}$}\\
\cline{1-7}
BM3D &31.08 &28.56 & 25.62 &\cellcolor[gray]{0.70}\color{blue}2 & \cellcolor[gray]{0.90}31.14 & \cellcolor[gray]{0.90}31.30\\
LSSC&31.27 &28.70 & 25.72 &\cellcolor[gray]{0.70}\color{blue}5 & 
\cellcolor[gray]{0.90}31.30 & \cellcolor[gray]{0.90}\textbf{31.42}\\
EPLL-GMM&31.19 &28.68 & 25.67 &\cellcolor[gray]{0.70}\color{blue}8 & 
\cellcolor[gray]{0.90}31.34 & \cellcolor[gray]{0.90}\textbf{31.43}\\
\cline{5-7}
$\text{opt-MRF}_{7 \times 7}$ &31.18 &28.66 & 25.70 & & 
\multicolumn{2}{c}{$\sigma = 25$}\\
$\text{RTF}^5$ &-- &28.75 & -- & \cellcolor[gray]{0.70}\color{blue}2 & \cellcolor[gray]{0.90}28.58& \cellcolor[gray]{0.90}28.77\\
WNNM&{31.37} &{28.83} & 25.83 & \cellcolor[gray]{0.70}\color{blue}5 & \cellcolor[gray]{0.90}28.78 & 
\cellcolor[gray]{0.90}\textbf{28.92}\\
$\text{CSF}_{5 \times 5}^5$&31.14 &28.60 & -- & \cellcolor[gray]{0.70}\color{blue}8 
& \cellcolor[gray]{0.90}{28.83} & \cellcolor[gray]{0.90}\textbf{28.95}\\
\cline{5-7}
$\text{CSF}_{7 \times 7}^5$ &31.24 &28.72 & -- & & \multicolumn{2}{c}{$\sigma = 50$}\\
$\text{ARF}_{5 \times 5}^4$ & 30.70 & 28.20 & -- & 
\cellcolor[gray]{0.70}\color{blue}2 & \cellcolor[gray]{0.90} 25.54 & \cellcolor[gray]{0.90} 25.78\\
$\text{opt-GD}_{5 \times 5}^{10}$ &-- & 28.39 &-- & 
\cellcolor[gray]{0.70}\color{blue}5 & \cellcolor[gray]{0.90} 25.80 & \cellcolor[gray]{0.90} \textbf{25.96}\\
\tikzmark{p2} &  & \tikzmark{b2} & \tikzmark{a2}& 
\cellcolor[gray]{0.70}\color{blue}8 & \cellcolor[gray]{0.90} \textbf{25.87} & \cellcolor[gray]{0.90} \textbf {26.01}\\
\Xhline{0.5pt}
\cline{1-7}
\connect{p2}{a2}
\end{tabular}
\caption{Average PSNR (dB) on 68 images from \cite{RothFOE2009} for image 
denoising with $\sigma = 15, 25 ~\text{and} ~50$. {All the TNRD models are 
jointly trained. 
Note that among those algorithms similar to our model, 
$\text{opt-MRF}_{7 \times 7}$, 
$\text{ARF}_{5 \times 5}^4$ and $\text{opt-GD}_{5 \times 5}^{10}$ only train the 
filters with fixed penalty function $\text{log}(1+z^2)$. In the 
$\text{opt-MRF}_{7 \times 7}$ model, 48 filters of size $7 \times 7$ 
(\textbf{2304 free parameters}), 
for $\text{ARF}_{5 \times 5}^4$, 13 filters of size $5 \times 5$ 
(\textbf{325 free parameters}) and 
for the $\text{opt-GD}_{5 \times 5}^{10}$ algorithm, 24 filters of size 
$5 \times 5$ (\textbf{600 free parameters}) are trained. 
The CSF model and our approach train both the filters and nonlinearities, thus 
involving more parameters, \eg, 
the $\text{TNRD}_{7\times 7}^5$ model involves \textbf{26,645 free parameters} and the 
corresponding 
the $\text{CSF}_{7\times 7}^5$ model involves \textbf{24,245 free parameters}.
}}\label{denoisingresults}
\vspace*{-0.75cm}
\end{table}

\subsection{Run time} 
The algorithm structure of our TNRD model is similar to the CSF model, which is well-suited 
for parallel computation on GPUs. We implemented our trained models on GPU using CUDA programming to 
speed up the inference procedure, and finally it leads to significantly improved 
performance, see Table \ref{run time}. 
We make a run time comparison to other denoising algorithms based on strictly enforced single-threaded CPU computation (
\eg, start Matlab with -singleCompThread) for a fair comparison, see Table \ref{run time}. We only 
present the results of some selective algorithms, which either have the best denoising result or run time performance. 
We refer to \cite{CSF2014} for a comprehensive run time comparison of various algorithms\footnote{
LSSC, EPLL, opt-MRF and $\text{RTF}^5$ methods are much slower than BM3D on the CPU, \textit{cf.} \cite{CSF2014}. }.

We see that our TNRD model is generally faster than the CSF model with the same model capacity. 
It is reasonable, because in each stage the CSF model involves additional DFT and inverse DTF operations, \ie, 
our model only requires a portion of the computation of the CSF model. Even though the BM3D is a non-local model, 
it still possesses high computational efficiency. In contrast, another non-local model - WNNM achieves compelling 
denoising results at the expense of huge computation time. Moreover, the WNNM algorithm is hardly applicable for 
high resolution images (\eg, 10 mega-pixels) due to its huge memory requirements. 
Note that our model can be also easily implemented with multi-threaded CPU computation. 

In summary, our $\text{TNRD}^5_{7\times 7}$ model outperforms these recent state-of-the-arts, 
meanwhile it is the fastest method even with a CPU implementation. 
We present an illustrative denoising example in Figure \ref{Gaussian25-2} on an image from the test dataset. 
More denoising examples can be found in the \textit{supplemental material} based on images from the test dataset 
and a megapixel-size natural image of size $1050 \times 1680$. 
\begin{table}[t!]
\centering
\hspace*{-0.2cm} \begin{tabular}{c c c c c c}
\Xhline{0.5pt}
\cline{1-6}
Method & $256^2$ & $512^2$ & $1024^2$ & $2048^2$ & $3072^2$\\
\cline{1-6}
\rowcolor[gray]{0.85} BM3D \cite{BM3D} &1.1 &4.0 &17 & 76.4 & {\color{black}{176.0}}\\
\rowcolor[gray]{0.85} $\text{CSF}_{7 \times 7}^5$ \cite{CSF2014} &3.27 &11.6 &40.82 & 151.2 & {\color{black}{494.8}}\\
\rowcolor[gray]{0.85} WNNM \cite{WNNM} &122.9 & 532.9 &2094.6 & -- & --\\
\cline{1-6}
\multirow{3}*{$\text{TNRD}^5_{5 \times 5}$} & \cellcolor[gray]{0.85} 0.51 & 
\cellcolor[gray]{0.85} 1.53 & \cellcolor[gray]{0.85} 5.48 
&\cellcolor[gray]{0.85} 24.97 & \cellcolor[gray]{0.85} 53.3\\
& {\color{blue}{0.43}}& {\color{blue}{0.78}}& {\color{blue}{2.25}}& {\color{blue}{8.01}}& {\color{blue}{21.6}}\\
& {\color{red}{0.005}} & {\color{red}{0.015}} & {\color{red}{0.054}} &{\color{red}{0.18}} & {\color{red}{0.39}}\\
\cline{1-6}
\multirow{3}*{$\text{TNRD}^5_{7 \times 7}$} & \cellcolor[gray]{0.85} 1.21 & 
\cellcolor[gray]{0.85} 3.72 & \cellcolor[gray]{0.85} 14.0 
&\cellcolor[gray]{0.85} 62.2& \cellcolor[gray]{0.85} 135.9\\
& {\color{blue}{0.56}}& {\color{blue}{1.17}}& {\color{blue}{3.64}}& {\color{blue}{13.01}}& {\color{blue}{30.1}}\\
& {\color{red}{0.01}} & {\color{red}{0.032}} &{\color{red}{0.116}} & {\color{red}{0.40}} & {\color{red}{0.87}}\\
\Xhline{0.5pt}
\cline{1-6}
\end{tabular}
\vspace*{0.2cm}
\caption{Run time comparison for image denoising (in seconds) with different implementations. 
(1) The run time results with \colorbox{grayB}{gray}background are evaluated 
with the single-threaded implementation on Intel(R) Xeon(R) CPU E5-2680 v2 @ 2.80GHz; 
(2) the {\color{blue}{blue}} colored run times are obtained with multi-threaded computation using Matlab \textit{parfor} 
on the above CPUs;
(3) the run time results colored in {\color{red}{red}} are executed on a NVIDIA GeForce GTX 780Ti GPU. 
We do not count the memory transfer time between CPU/GPU for the GPU implementation (if counted, the run time will nearly 
double)}\label{run time}
\vspace*{-1cm}\end{table}

\begin{figure*}[t!]
\centering
    \subfigure[Noisy, 20.17dB]{\includegraphics[width=0.325\linewidth]{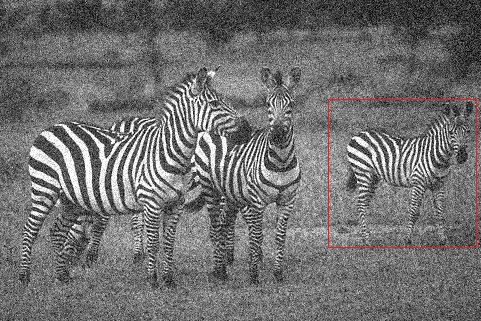}}\hfill
    \subfigure[BM3D, 27.53dB/\textbf{CPU: 2.5s}]{\includegraphics[width=0.325\linewidth]{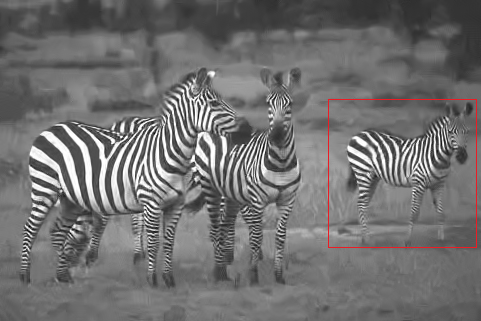}}\hfill
    \subfigure[$\text{CSF}_{7 \times 7}^5$, 28.00dB/\textbf{GPU: 0.55s}]
{\includegraphics[width=0.325\linewidth]{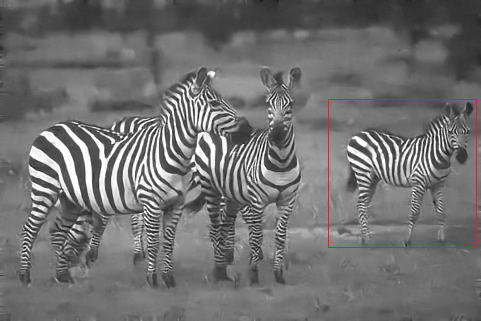}}\\
\vspace*{-0.15cm}
    \subfigure[WNNM, 27.94dB/\textbf{CPU: 393.2s}]{\includegraphics[width=0.325\linewidth]{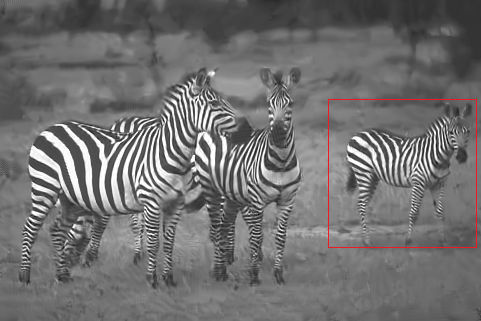}}\hfill
    \subfigure[$\text{TNRD}_{5 \times 5}^5$, 28.16dB/\textbf{GPU: 9.1ms}]
{\includegraphics[width=0.325\linewidth]{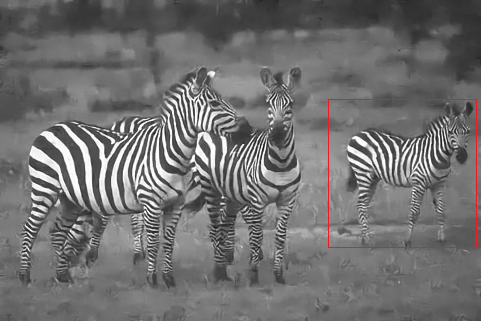}}\hfill
    \subfigure[$\text{TNRD}_{7 \times 7}^5$, {\color{red}{28.23dB}}/\textbf{GPU: 20.3ms}]{\includegraphics[width=0.325\linewidth]
{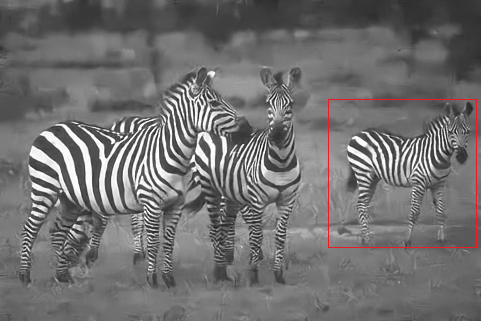}}\\
    \caption{Denoising results on a test image of size $481 \times 321$ ($\sigma = 25$) by different methods 
(compared with BM3D \cite{BM3D}, WNNM \cite{WNNM} and CSF model \cite{CSF2014}), together with the 
corresponding computation time either on CPU or GPU. 
Note the differences in the highlighted region.}\label{Gaussian25-2}
\end{figure*}

\section{Single image super resolution (SISR)}
\begin{table*}[t!]
\centering
\begin{tabular}{|l| c| c| c| c| c| c| c| c| c| c| c|c| c|}
\hline
\textbf{Set5} & & \multicolumn{2}{c|}{Bicubic}& \multicolumn{2}{c|}{K-SVD \cite{SRKSVD}}& \multicolumn{2}{c|}{ANR \cite{ANR}}
& \multicolumn{2}{c|}{SR-CNN \cite{SRCNN}} & \multicolumn{2}{c|}{RFL \cite{schulter15}} 
& \multicolumn{2}{c|}{$\text{TNRD}_{7 \times 7}^5$} \\
\cline{3-14}
images & scale & PSNR & Time& PSNR & Time& PSNR & Time& PSNR & Time& PSNR & Time& PSNR & Time\\
\hline\hline
baby & 2 & 37.07 & - &38.25 & 8.21 & 38.44 & 1.39 & 38.30 & 0.38 & 38.39 & 1.31 & \textbf{38.51} & 1.52\\
bird  & 2 & 36.81 & - &39.93 & 2.67 & 40.04 & 0.44 & 40.64 & 0.14 & 40.99 & 0.52 & \textbf{41.29} & 0.59 \\
butterfly  & 2 & 27.43 & - & 30.65 & 2.14 & 30.48 & 0.38 & 32.20 & 0.10 & 32.46 & 0.41 & \textbf{33.16} & 0.56\\
head  & 2 & 34.86 & - & 35.59 & 2.46 &35.66 & 0.41 & 35.64 & 0.13 & \textbf{35.70} & 0.48 & \textbf{35.71} & 0.60\\
woman  & 2 & 32.14 & - & 34.49 & 2.45 & 34.55 & 0.43 & 34.94 & 0.13 & 35.19 & 0.46 & \textbf{35.50} & 0.57\\
\hline\hline
\rowcolor[gray]{0.85} \textbf{average}  & 2 
& 33.66 & - & 35.78 & 3.59 & 35.83 & 0.61 & 36.34 & 0.18 & 36.55 & 0.64 & \textbf{36.83} & 0.77\\
\hline\hline
baby & 3 & 33.91 & - &35.08 & 3.77 & 35.13 & 0.79 & 35.01 & 0.38 & 35.04 & 0.79 & \textbf{35.28} & 1.52\\
bird  & 3 & 32.58 & - &34.57 & 1.34 & 34.60 & 0.27 & 34.91 & 0.14 & 35.15 & 0.31 & \textbf{36.11} & 0.59 \\
butterfly  & 3 & 24.04 & - & 25.94 & 1.08 & 25.90 & 0.24 & 27.58 & 0.10 & 27.18 & 0.25 & \textbf{28.90} & 0.56\\
head  & 3 & 32.88 & - & 33.56 & 1.35 &33.63 & 0.24 & 33.55 & 0.13 & 33.68 & 0.29 & \textbf{33.78} & 0.60\\
woman  & 3 & 28.56 & - & 30.37 & 1.14 & 30.33 & 0.24 & 30.92 & 0.13 & 30.92 & 0.28 & \textbf{31.77} & 0.57\\
\hline\hline
\rowcolor[gray]{0.85} \textbf{average}  & 3
& 30.39 & - & 31.90 & 1.74 & 31.92 & 0.35 & 32.39 & 0.18 & 32.39 & 0.39 & \textbf{33.17} & 0.77\\
\hline\hline
baby & 4 & 31.78 & - & 33.06 & 2.63 & 33.03 & 0.59 & 32.98 & 0.38 & 33.05 & 0.60 & \textbf{33.29} & 1.52\\
bird  & 4 & 30.18 & - &31.71 & 0.70 & 31.82 & 0.18 & 31.98 & 0.14 & 32.14 & 0.23 & \textbf{32.98} & 0.59 \\
butterfly  & 4 & 22.10 & - & 23.57 & 0.54 & 23.52 & 0.14 & 25.07 & 0.10 & 24.44 & 0.19 & \textbf{26.22} & 0.56\\
head  & 4 & 31.59 & - & 32.21 & 0.66 &32.27 & 0.16 & 32.19 & 0.13 & 32.31 & 0.22 & \textbf{32.57} & 0.60\\
woman  & 4 & 26.46 & - & 27.89 & 0.72 & 27.80 & 0.23 & 28.21 & 0.13 & 28.31 & 0.23 & \textbf{29.17} & 0.57\\
\hline\hline
\rowcolor[gray]{0.85} \textbf{average}  & 4
& 28.42 & - & 29.69 & 1.05 & 29.69 & 0.26 & 30.09 & 0.18 & 30.05 & 0.29 & \textbf{30.85} & 0.77\\
\hline
\end{tabular}\vspace*{0.2cm}
\caption{PSNR (dB) and run time (s) performance for upscaling factors $\times 2$, $\times 3$ and $\times 4$ on the 
\textbf{Set5} dataset. All the methods use the same 91 training images as in \cite{ANR}.}\label{table:set5}
\vspace*{-0.5cm}
\end{table*}

\begin{table*}[t!]
\centering
\begin{tabular}{|l| c| c| c| c| c| c| c| c| c| c|c| c|}
\hline
\textbf{Set14} & \multicolumn{2}{c|}{Bicubic}& \multicolumn{2}{c|}{K-SVD \cite{SRKSVD}}& \multicolumn{2}{c|}{ANR \cite{ANR}}
& \multicolumn{2}{c|}{SR-CNN \cite{SRCNN}} & \multicolumn{2}{c|}{RFL \cite{schulter15}} 
& \multicolumn{2}{c|}{$\text{TNRD}_{7 \times 7}^5$} \\
\cline{2-13}
images& PSNR & Time& PSNR & Time& PSNR & Time& PSNR & Time& PSNR & Time& PSNR & Time\\
\hline\hline
baboon & 23.21 & - &23.52 & 3.54 & 23.56 & 0.77 & 23.60 & 0.40 & 23.57 & 0.75 & \textbf{23.62} & 1.30\\
barbara & 26.25 & - &\textbf{26.76} & 6.24 & 26.69 & 1.23 & 26.66 & 0.70 & 26.63 & 1.18 & 26.25 & 1.75 \\
bridge & 24.40 & - & 25.02 & 3.98 & 25.01 & 0.80 & 25.07 & 0.44 & 25.11 & 0.81 & \textbf{25.29} & 1.19\\
coastguard & 26.55 & - & 27.15 & 1.54 &27.08 & 0.36 & \textbf{27.20} & 0.17 & 27.16 & 0.35 & 27.12 & 0.65\\
comic & 23.12 & - & 23.96 & 1.37 & 24.04 & 0.27 & 24.39 & 0.15 & 24.27 & 0.34 & \textbf{24.67} & 0.65\\
face & 32.82 & - & 33.53 & 1.10 & 33.62 & 0.24 & 33.58 & 0.13 & 33.65 & 0.29 & \textbf{33.82} & 0.57\\
flowers & 27.23 & - & 28.43 & 2.66 &28.49 & 0.57 & 28.97 & 0.30 & 28.86 & 0.61 & \textbf{29.55} & 0.90\\
foreman & 31.18 & - & 33.19 & 1.54 & 33.23 & 0.30 & 33.35 & 0.17 & 33.87 & 0.36 & \textbf{34.65} & 0.65\\
lenna & 31.68 & - & 33.00 & 3.89 & 33.08 & 0.79 & 33.39 & 0.44 & 33.38 & 0.77 & \textbf{33.77 } & 1.19 \\
man & 27.01 & - & 27.90 & 3.81 & 27.92 & 0.76 & 28.18 & 0.44 & 28.20 & 0.80 & \textbf{28.52} & 1.17\\
monarch & 29.43 & - & 31.10 & 6.13 &31.09 & 1.13 & 32.39 & 0.66 & 32.10 & 1.12 & \textbf{33.61} & 1.66\\
pepper & 32.39 & - & 34.07 & 3.84 & 33.82 & 0.80 & 34.35 & 0.44 & 34.55 & 0.82 & \textbf{35.06} & 1.20\\
ppt3 & 23.71 & - & 25.23 & 4.53 & 25.03 & 1.01 & 26.02 & 0.58 & 25.84 & 0.98 & \textbf{27.08} & 1.48\\
zebra & 26.63 & - & 28.49 & 3.36 & 28.43 & 0.69 & 28.87 & 0.38 & 29.03 & 0.72 & \textbf{29.40} & 1.04\\
\hline\hline
\rowcolor[gray]{0.85} \textbf{average performance} 
& 27.54 & - & 28.67 & 3.40 & 28.65 & 0.69 & 29.00 & 0.39 & 29.02 & 0.71 & \textbf{29.46} & 1.10\\
\hline
\end{tabular}\vspace*{0.2cm}
\caption{Upscaling factor $\times 3$ performance in terms of PSNR(dB) and runtime (s) per image on the 
\textbf{Set14} dataset.}\label{table:set14}
\vspace*{-0.75cm}
\end{table*}

\begin{figure*}[t!]
\centering
    \subfigure[Bicubic / 29.43dB]{\includegraphics[width=0.325\linewidth]{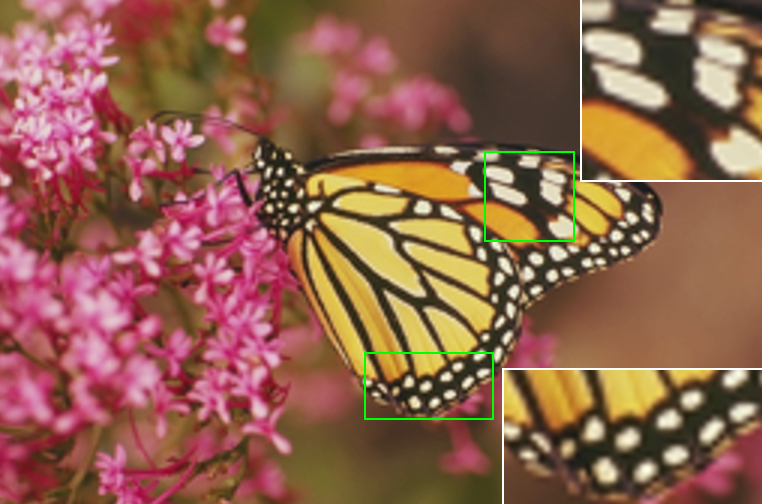}}\hfill
    \subfigure[K-SVD \cite{SRKSVD} / 31.10dB]{\includegraphics[width=0.325\linewidth]{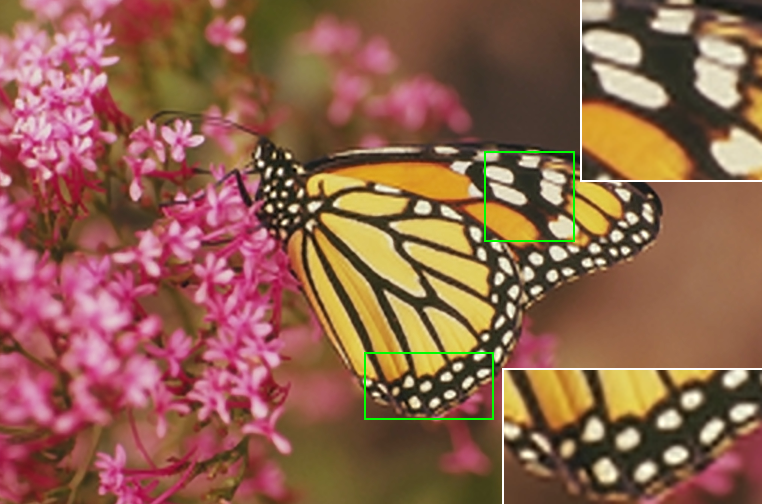}}\hfill
    \subfigure[ANR \cite{ANR} / 31.09dB]{\includegraphics[width=0.325\linewidth]{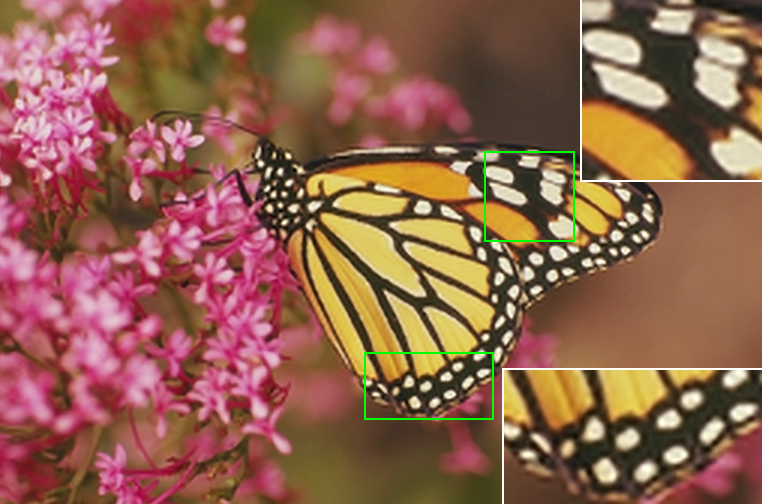}}\\
\vspace*{-0.15cm}
    \subfigure[SR-CNN \cite{SRCNN} / 32.39dB]{\includegraphics[width=0.325\linewidth]{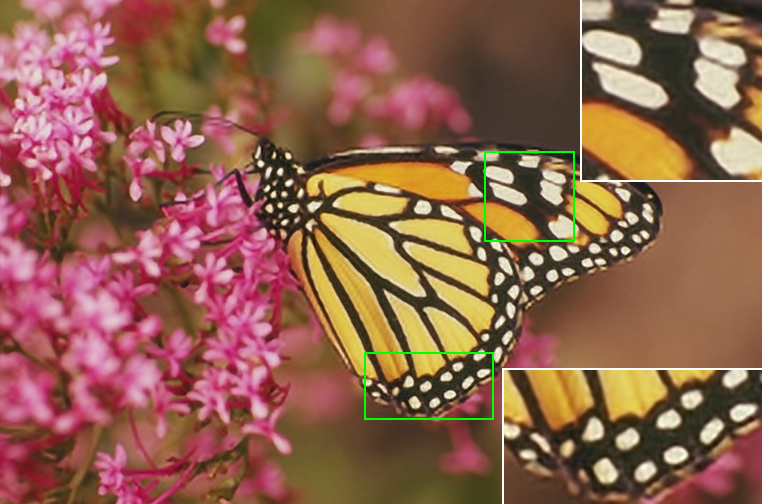}}\hfill
    \subfigure[RFL \cite{schulter15} / 32.10dB]{\includegraphics[width=0.325\linewidth]{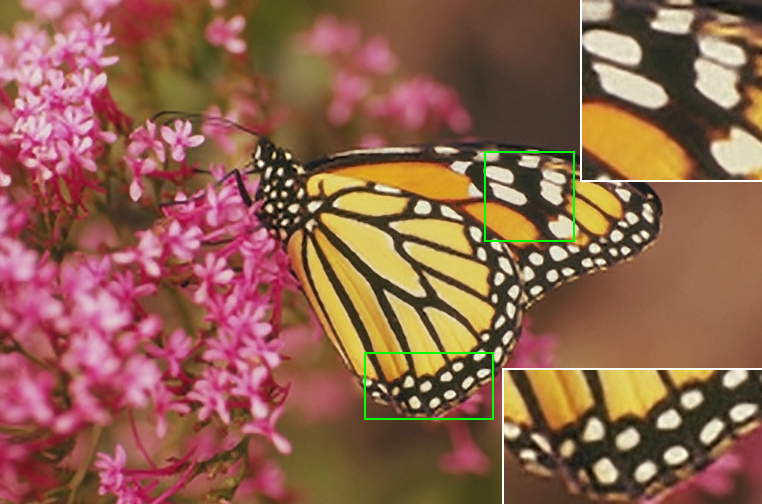}}\hfill
    \subfigure[$\text{TNRD}_{7 \times 7}^5$ / {\color{red}{33.61dB}}]{\includegraphics[width=0.325\linewidth]
{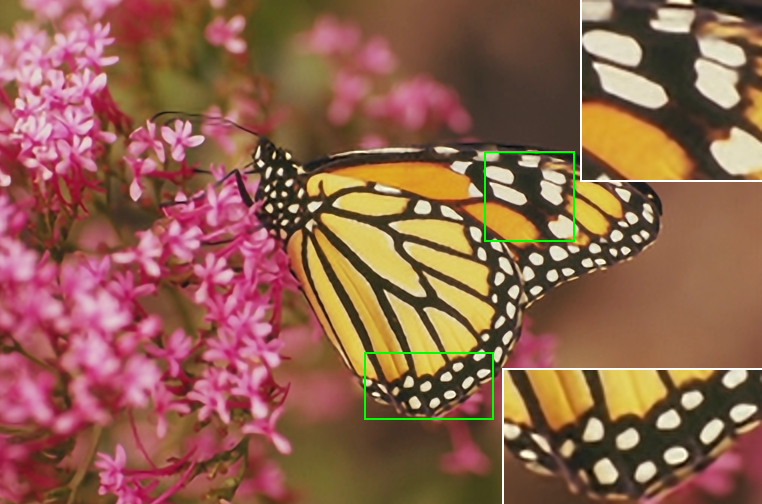}}\\
\vspace*{-0.25cm}
    \caption{A super resolution example for the ``Monarch'' image from \textbf{Set14} 
with an upscaling factor $\times 3$. Note the differences in the highlighted region that 
our model achieves more clean and sharp image edges. \textbf{Best viewed on screen and zoom in.}}\label{fig:SR}
\vspace*{-0.25cm}
\end{figure*}

As demonstrated in the last section that our trained diffusion model can lead to explicit backward diffusion process, which 
sharpens image structures like edges. This is the very property demanded for the task of image super resolution. Therefore, 
we are motivated to investigate the SISR problem with our proposed approach. 

We start with the following energy functional
\begin{equation}\label{vmsr}
\min\limits_{u}E(u) = 
\suml{i=1}{N_k}\rho_i(k_i * u) + \frac{\lambda}{2}\|Au - f\|_2^2\,,
\end{equation}
where the linear operator $A$ is a bicubic interpolation which 
links the high resolution (HR) image $h$ to the low resolution (LR) image $f$ via 
$f = Ah$. 
Casting $\cD(u) = \frac{\lambda}{2}\|Au - f\|^2_2$ and $\cG(u)  = 0$, 
the energy functional \eqref{vmsr} suggests the following diffusion process 
\begin{equation}\label{SRdiffusion}
u_t = u_{t-1} - \left(\sum\limits_{i = 1}^{N_k}\bar k_i^t * \phi_i^t(k_i^t * u_{t-1}) + \lambda^t A^\top (A u_{t-1} - f)
\right)\,, 
\end{equation}
where the starting point $u_0$ 
is given by the direct bicubic interpolation of the LR image $f$. 
Computing the gradients $\frac{\partial u_t}{\partial \Theta_t}$ and $\frac{\partial u_t}{\partial u_{t-1}}$ with respect to 
\eqref{SRdiffusion} can be done with little modifications to the derivations for image denoising. 
Detailed derivations are presented in the \textit{supplemental material}. 

We considered the model capacity of $\text{TNRD}_{7 \times 7}^5$, and trained diffusion models for three upscaling 
factors $\times 2$, $\times 3$ and $\times 4$, using exactly 
the same 91 training images as in previous works \cite{ANR, schulter15}. The trained models are evaluated on two 
different test data sets: \textbf{Set5} and \textbf{Set14}. Following previous works \cite{ANR, SRCNN, schulter15}, the trained 
models are only applied to the luminance component of an image, and a regular bicubic upscaling method is applied to 
the color components. 

The test results are summarized in Table \ref{table:set5} and 
Table \ref{table:set14}. One can see that in terms of average PSNR, 
our trained diffusion model $\text{TNRD}_{7 \times 7}^5$ leads to significant 
improvements over very recent state-of-the-arts in all cases, meanwhile it is still among the fast algorithms\footnote{Note that our approach is a Matlab implementation, while some of other algorithms are 
based on C++ implementations, such as SR-CNN.}. A SISR example is shown in Figure \ref{fig:SR} 
to illustrate its effectiveness. One can see that our approach can obtain more 
accurate image edges, 
as shown in the zoom-in parts. More SISR examples can be found in the \textit{supplemental material}. 

\begin{figure*}[t!]
\centering
    \subfigure[Original]{\includegraphics[width=0.16\linewidth]{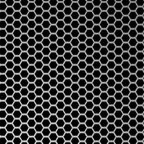}}\hfill
    \subfigure[Bicubic / 14.18dB]{\includegraphics[width=0.16\linewidth]{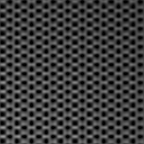}}\hfill
    \subfigure[ANR / 15.07dB]{\includegraphics[width=0.16\linewidth]{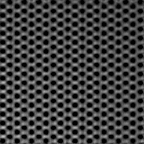}}\hfill
    \subfigure[SR-CNN / 15.65dB]{\includegraphics[width=0.16\linewidth]{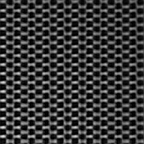}}\hfill
    \subfigure[RFL / 15.03dB]{\includegraphics[width=0.16\linewidth]{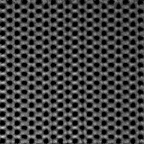}}\hfill
    \subfigure[$\text{TNRD}_{7 \times 7}^5$ / {{17.22dB}}]{\includegraphics[width=0.16\linewidth]
{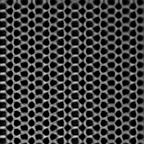}}\\
\vspace*{-0.25cm}
    \caption{A toy experiment on a synthesized image with repeated hexagons for the upscaling factor $\times 3$.}\label{fig:SRtoy}
\end{figure*}

We apply the learned diffusion parameters to the diffusion equation \eqref{diffusion}. It turns out that 
the diffusion process can also generate some interesting patterns from random images, as shown in Figure \ref{patterns}. 
We believe that this 
ability to generate image patterns from weak evidence is the main reason for the superiority of our trained model for the 
SISR task. In order to further validate our argument, we carry out a toy SISR experiment based on a synthesized image 
with repeated hexagons. The results are shown in Figure \ref{fig:SRtoy}, where one can see that our trained model can better 
reconstruct those repeated image structures. 
\vspace*{-0.3cm}
\section{JPEG deblocking experiments}\label{JPEG}
In order to further demonstrate the applicability of our proposed framework for those problems with a non-smooth data term, 
we investigate the JPEG deblocking problem - suppressing the block artifacts in the JPEG compressed images, which is 
formulated as a non-smooth optimization problem. 
Motivated by \cite{BrediesH12}, we consider the following variational model based on the FoE image prior 
\begin{equation}\label{vbdeblocking}
\min\limits_{u}E(u) = 
\suml{i=1}{N_k}\rho_i(k_i * u) + \cI_{Q} (Du)\,,
\end{equation}
where $\cI_{Q}$ is a indicator function over the set $Q$ (quantization constraint set). In JPEG compression, information loss 
happens in the quantization step, where all possible values in the range $[d - 0.5, d + 0.5]$ ($d$ is an integer) are quantized to a 
single number $d$. Given a compressed data, we only know $d$. Therefore, all possible values in the interval $[d - 0.5, d + 0.5]$ 
define a convex set $Q$ which is a box constraint. The sparse matrix 
$D \in \R^{N \times N}$ denotes the block DCT transform. We refer to \cite{BrediesH12} for more details. 

By setting $\cD(u) = 0$ and $\cG(u)  = \cI_{Q} (Du)$, we obtain the following 
diffusion process
\begin{equation}\label{deblocking}
u_t = D^\top \text{proj}_{Q}\left(D \left(u_{t-1} - \sum\nolimits_{i = 1}^{N_k}\bar k_i^t * \phi_i^t(k_i^t * u_{t-1})\right) \right) \,,
\end{equation}
where $\text{proj}_{Q}(\cdot)$ denotes the orthogonal projection onto $Q$. More details can be found in the 
\textit{supplemental material}. 

We also trained diffusion models for the JPEG deblocking problem. 
We followed the test procedure in \cite{ECCV2012RTF} for performance evaluation. 
We distorted the images by JPEG blocking artifacts. We considered 
three compression quality settings $q = 10, 20 ~\text{and} ~30$ for the JPEG encoder. 

We trained three nonlinear diffusion $\text{TNRD}_{7 \times 7}$ models for different compression parameter $q$. 
We found that for JPEG deblocking, 4 stages are already enough. Results of the trained models are shown in Table 
\ref{results}, compared with several representative deblocking approaches. 
We see that our trained $\text{TNRD}^4_{7 \times 7}$ outperforms all the 
competing approaches in terms of PSNR. {
Concerning the run time, our model takes about 11.2s 
to handle an image of size $1024 \times 1024$ with CPU computation, 
while the strongest competitor (in terms of run time) - SADCT 
consumes about 56.5s\footnote{
RTF is slower than SADCT, as it depends on the output of SADCT.}. 
Furthermore, our model is extremely fast on GPUs. 
For the same image size the GPU implementation takes about 0.095s.} 
See the \textit{supplemental material} for JPEG deblocking examples.
\begin{table}[t!]
\vspace*{0.25cm}
\centering
\hspace*{-0.15cm} \begin{tabular}{|l|L{0.8cm}|C{0.8cm}|C{0.9cm}|C{0.8cm}|C{0.9cm}|C{1.2cm}|}
\hline
$q$ & {\small JPEG \newline decoder} & \small TGV \newline \cite{BrediesH12}& \small Dic-SR\cite{TVdeblockingDic}
&\footnotesize SADCT \cite{foi2007pointwise} &\small RTF\cite{ECCV2012RTF} & \small $\text{TNRD}^4_{7 \times 7}$\\
\hline\hline
10 & 26.59 & 26.96 & 27.15 & 27.43 & {27.68} & \textbf{27.85}\\
\hline
20 & 28.77 & 29.01 & 29.03 & 29.46 & {29.83} & \textbf{30.06}\\
\hline
30 &  30.05 & 30.25 & 30.13 & 30.67 & {31.14} & \textbf{31.41}\\
\hline
\end{tabular}
\vspace*{0.2cm}
\caption{JPEG deblocking results for natural images, reported with average PSNR values.}\label{results}
\vspace*{-0.75cm}\end{table}
\section{Discussion, summary and future work}
\subsection{Summary}
We have proposed a trainable nonlinear reaction diffusion framework for effective image restoration. Its critical point lies in 
the additional training of the influence functions. We have trained our models for 
the problem of Gaussian denoising, single image super resolution and JPEG deblocking. Based on standard test datasets, 
the trained models result in the best-reported results. 
We believe that the effectiveness of the trained diffusion models is attributed to the following desired properties of the models
\begin{itemize}
\setlength\itemsep{0em}
    \item \noindent \textit{Anisotropy}. In the trained filters, we can find rotated derivative filters in different directions, \textit{cf.} Fig 
\ref{fig:filters}, which will make the diffusion happen in some special directions. 
    \item \textit{Higher order}. The learned filters contain first, second and higher-order derivative filters, \textit{cf.} Fig 
\ref{fig:filters}.  
    \item \textit{Adaptive forward/backward diffusion through the learned nonlinear functions}. Nonlinear functions corresponding to 
explicit backward diffusion appear in the learned nonlinearity, \textit{cf.} Fig \ref{functions}.
\end{itemize}

Meanwhile, the structure of trained models is very simple 
and well-suited for parallel computation on GPUs. As a consequence, 
the resulting algorithms are significantly faster than all competing algorithms and hence are also applicable to the restoration 
of high resolution images. 

\subsection{Discussion}
One possible limitation of the proposed TNRD approach is that one has to define the ground truth - the expected output of the diffusion network during training. 
For image restoration applications in this paper, this is not a problem as we have a clear choice for the ground truth. 
However, for those applications with ambiguous ground truth, \eg, image structure extraction \cite{xu2012structure}, 
we will have to make efforts to define the ground truth.

Furthermore, the trained diffusion networks will only perform well in the way they are trained. For example, the trained model 
based on noise level $\sigma = 25$ will break for an input image with noise $\sigma = 50$, and the trained model for 
upscaling factor $\times 3$ will also lead to inferior performance when it is applied to the SISR problem of 
upscaling factor $\times 2$. It is generally hard to train a universal diffusion model to handle all the noise levels or 
all upscaling factors. 

Our approach is to optimize a time-discrete PDE, which is inspired by 
FoE based model, but we do not aim to minimize 
a series of FoE based energies. Our model directly learns an optimal trajectory 
for a certain possibly unknown energy functional, 
the minimizer of which provides a good estimate of the demanded solution. 
Probably, such a functional can not be modeled by a single FoE energy, 
while our learned gradient descent/forward-backward 
steps provide good approximation to the local gradients of this unknown functional.

\subsection{Future work} 
From an application point of view, we
think that it will be interesting to consider learned nonlinear reaction diffusion based models also 
for other image processing tasks
such as image inpainting, blind image deconvolution, optical
flow. Moreover, since learning the influence functions turned out to
be crucial, we believe that learning optimal nonlinearities (\ie, activation functions) in standard CNs
could lead to a similar performance increase. There are actually two recent works \cite{agostinelli2014learning, he2015delving} 
to investigate a parameterized ReLU function in standard deep convolutional networks, which indeed brings improvements even 
with little freedom to tune the activation functions. Finally, it will also be
interesting to investigate the unconventional penalty functions
learned by our approach in usual energy minimization approaches.

{\footnotesize
\bibliographystyle{ieee}
\bibliography{cvpr_bib}
}

\begin{IEEEbiography}[{\includegraphics[width=1in,height=1.25in,clip,keepaspectratio]
{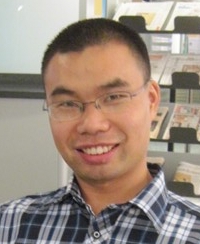}}]{Yunjin Chen}
received a B.Sc. degree in applied physics from Nanjing University of Aeronautics 
and Astronautics, China, a M.Sc. degree in optical engineering from National 
University of Defense Technology, China, and a Ph.D degree in computer science from 
Graz University of Technology, Austria in 2007, 2010 and 2015, 
respectively. Since July 2015, he serves as 
a scientific researcher in the military of China. His current research interests are 
learning image prior model for low-level computer 
vision problems and convex optimization. 
\end{IEEEbiography}
\begin{IEEEbiography}[{\includegraphics[width=1in,height=1.25in,clip,keepaspectratio]
{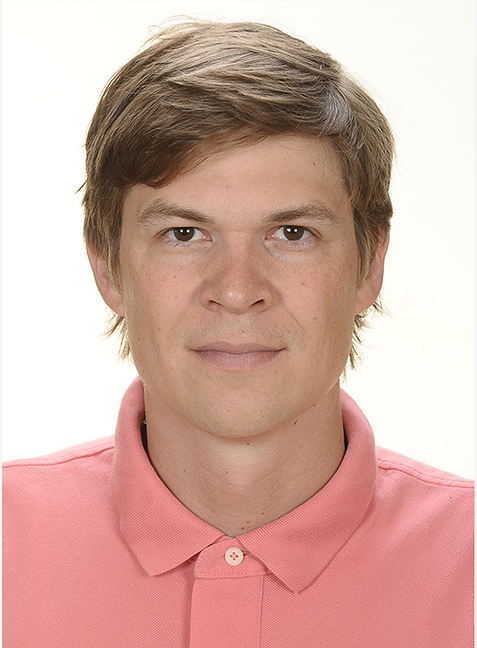}}]{Thomas Pock}
received a M.Sc and a Ph.D degree in Computer Engineering 
from Graz University of Technology in 2004 and 2008, respectively. 
In 2013 he received the START price of the Austrian Science Fund (FWF) 
and the German Pattern recognition award of the German association for 
pattern recognition (DAGM). 
Since June 2014, he is a Professor of Computer Science at Graz University of 
Technology (AIT Stiftungsprofessur "Mobile Computer Vision") 
and a principal scientist at the Department of Safety and Security at the 
Austrian Institute of Technology (AIT). 
The focus of his research is the development of mathematical models for computer 
vision and image processing in mobile scenarios as well as the 
development of efficient algorithms to compute these models.
\end{IEEEbiography}

\end{document}